\documentclass[letterpaper, 10 pt, conference]{ieeeconf}  % Comment this line out if you need a4paper

\IEEEoverridecommandlockouts                              % This command is only needed if
\usepackage{amsmath} % assumes amsmath package installed
\usepackage{amssymb}  % assumes amsmath package installed
\usepackage{siunitx}
\usepackage{nicefrac}
\usepackage{paralist}
\usepackage{todonotes}
\usepackage{soul}
\usepackage{graphicx}
\usepackage{epstopdf}

\usepackage{pgfplots,lipsum}
\usepackage{booktabs}
\pgfplotsset{compat=newest}
\usepackage{scalefnt}
\usepackage{multirow}
\usepackage[font={small}]{caption}
\usepackage{subcaption}
\usepackage{wrapfig}
\usepackage[super]{nth}
\usepackage{fancyhdr}
\usepackage{pgfplotstable}
\usepackage{stfloats}
\usepackage{hyperref}
\newsavebox{\measurebox}

\title{\LARGE \bf
Courteous Behavior of Automated Vehicles at Unsignalized Intersections via Reinforcement Learning}

\author{Shengchao Yan, Tim Welschehold, Daniel B\"uscher, Wolfram Burgard% <-this % stops a space
\thanks{This project was funded through the Priority Programme “Cooperative Interacting Automobiles” of the German Science Foundation DFG.}% <-this % stops a space
\thanks{
All authors are with the Department of Computer Science, University of Freiburg, Germany.
        {\tt\small \{yan, twelsche, buescher, burgard\} @cs.uni-freiburg.de}}%
}

\begin{document}

\maketitle
\thispagestyle{empty}
\pagestyle{empty}

%%%%%%%%%%%%%%%%%%%%%%%%%%%%%%%%%%%%%%%%%%%%%%%%%%%%%%%%%%%%%%%%%%%%%%%%%%%%%%%%
\begin{abstract}
The transition from today's mostly human-driven traffic to a purely automated one will be a gradual evolution, with the effect that we will likely experience mixed traffic in the near future. Connected and automated vehicles can benefit human-driven ones and the whole traffic system in different ways, for example by improving collision avoidance and reducing traffic waves. Many studies have been carried out to improve intersection management, a significant bottleneck in traffic, with intelligent traffic signals or exclusively automated vehicles. However, the problem of how to improve mixed traffic at unsignalized intersections has received less attention. In this paper, we propose a novel approach to optimizing traffic flow at intersections in mixed traffic situations using deep reinforcement learning. Our reinforcement learning agent learns a policy for a centralized controller to let connected autonomous vehicles at unsignalized intersections give up their right of way and yield to other vehicles to optimize traffic flow. We implemented our approach and tested it in the traffic simulator SUMO based on simulated and real traffic data. The experimental evaluation demonstrates that our method significantly improves traffic flow through unsignalized intersections in mixed traffic settings and also provides better performance on a wide range of traffic situations compared to the state-of-the-art traffic signal controller for the corresponding signalized intersection. 
% The performance gain grows as more automated vehicles are employed. 
\end{abstract}

%%%%%%%%%%%%%%%%%%%%%%%%%%%%%%%%%%%%%%%%%%%%%%%%%%%%%%%%%%%%%%%%%%%%%%%%%%%%%%%%
\section{Introduction}
\label{sec:intro}
Over the past decades we observed a strong increase in the mobility of the population around the world. While, in general, this can be regarded as an indication of an improved quality of life, it does come with a strong increase in overall and individual traffic, creating a variety of problems, from increased travel duration and high energy consumption to high environmental pollution.
A promising and practical solutions to this problem is to make road traffic as efficiently as possible. In particular, since intersections represent one of the major bottlenecks of traffic flow~\cite{wu2015influence}, optimizing intersection management is currently highly important to improve traffic efficiency and safety.

In the past, traffic regulation relied on traffic polices, semaphores, traffic lights, traffic signs and sets of rules for intersection management. Furthermore, drivers also use turn signals, brake lights and even hand signals to communicate and cooperate with other traffic participants.
Traffic control signals are not panacea for intersection problems~\cite{MUTCD2009}. For example, they may reduce traffic efficiency for low or unbalanced traffic demand. Moreover, the control of every traffic light should be adjusted according to the traffic pattern of its location. Although recent works~\cite{varaiya2013max, 9340784} developed adaptive traffic signal control methods, for most intersections, which often have only one lane per road and mostly small traffic volume,
the use of static road signs assigning priority has proven to be more efficient~\cite{MUTCD2009}.

%Therefore we utilize sensors like rear-view camera to enhance drivers' perception capability, electromechanical systems like ABS/ESP\todo{cite? full name?} to compensate human's action accuracy and navigation devices to help human with planning and memorizing. However, these improvements are mostly focused on individual vehicles rather than the traffic system, such as vehicles' cooperation at intersections. 
%In the recent decades the remarkable advancements of computer science, wireless communication and machine perception have been accelerating the development of advanced driver-assistance systems (ADAS), vehicle to everything communication (V2X) and autonomous vehicles (AV). The promised land of traffic system is getting more and more attention, where the smart, fast, accurate and tireless computer drivers can do much better than their human predecessors by communicating with the environment and carrying out the optimal actions. Computers are better than humans at long-term memory and complex calculation, which are both essential for the optimization problem at intersections. How can we take advantage of the benefit brought by connected and autonomous vehicles (CAV) to improve IM? 

Nowadays, the first autonomous vehicles are mingling with the traffic and it is to be expected that their share will steadily increase in the future. Besides overcoming human limitations in driving, which are the main reason for accidents in traffic, these autonomous vehicles will supposedly be interconnected and thus offer new, more efficient ways of communication and traffic management.
Based on the expectation that future traffic will consist of connected autonomous vehicles (CAVs), a large majority of current research excludes human-driven vehicles (HVs) in their development of traffic management approaches. However, it might take decades for the technology, the infrastructure and the users to be ready for traffic with only connected autonomous vehicles~\cite{litman2020autonomous}. We therefore believe that, for the near future, applicable traffic management solutions must \textit{i)} consider various degrees of mixed traffic, \textit{ii)} pose no complications or major adjustment requests for human-driven vehicles, and \textit{iii)} not present a traffic disturbance or danger when the communication between the connected autonomous vehicles fails.

%Some works~\cite{varaiya2013max, 9340784} have been carried out to improve the performance of traffic lights, but it seems impossible to assist human drivers at unsignalized intersections to make better decisions, which are the vast majority of all the intersections~\cite{10.1145/1860058.1860077}.

\begin{figure}
    \centering
    \includegraphics[width=0.45\textwidth]{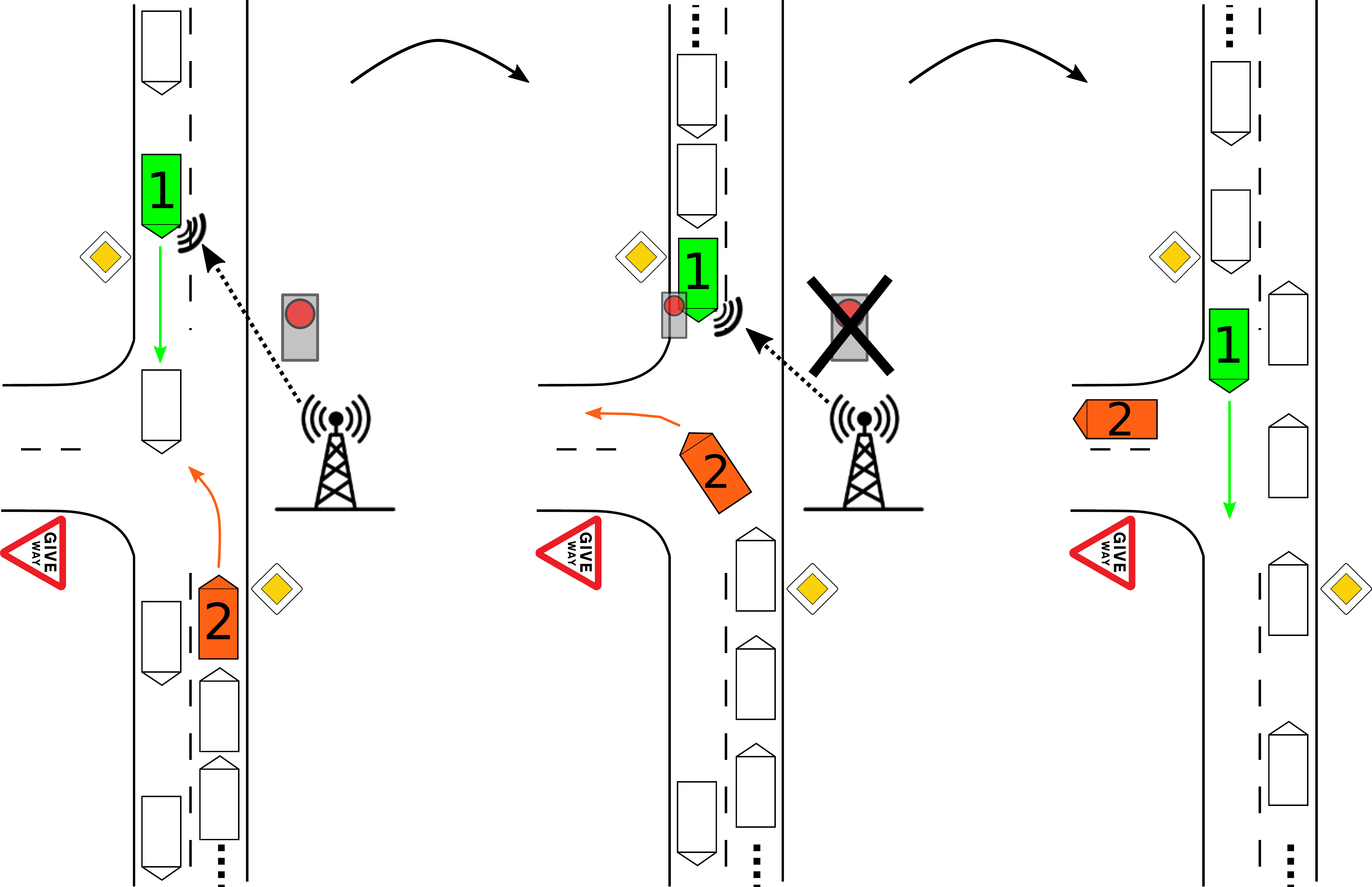}
    \caption{Our intersection management agent optimizes traffic flow by assigning virtual red traffic lights to connected autonomous vehicles (vehicle number 1). Once vehicle 2 is released, the vehicles following it can also proceed through the intersection.}
    \label{fig:introfig}
\end{figure}

One might argue that HVs lack means of efficient communication and coordination with other road users so that unsignalized intersections with mixed traffic cannot benefit from the introduction of CAVs~\cite{namazi2019intelligent}. However, Ulbrich \emph{et al.}~\cite{ulbrich2015structuring} showed that humans cooperate with other traffic participants to improve the whole traffic utility.
Consider, as an example, the situation shown in Fig.~\ref{fig:introfig}. Let us assume that vehicle 1 (green) is driven by a human. Even though it has higher priority and can proceed through the intersection before vehicle 2 (orange), the driver might prefer to yield to vehicle 2 so that the traffic behind vehicle 2 can be released sooner.
In general, such a decision is not trivial to make for a human because of several reasons.
First, due to occlusions the human driver might not see all upcoming vehicles and their turn signals, i.e., the driver in vehicle 1 might not have enough instantaneous information about the situation. 
Second, because of the short period of time elapsed since approaching the junction, the driver lacks relevant long term information, e.g., how long the vehicles with lower priority have been waiting. Third, although the information above could be provided using vehicle communication, it might distract the driver from the main driving task and thus pose a safety risk.
%\textit{Reaction time and driving habits:} Even installing a control unit that performs aforementioned perception and calculation to suggest a behavior to the HVs in our opinion is not a desirable solution. On the one hand this would still require taking into account humans' reaction to these extremely dynamic signals and might on the other hand confuse drivers as this would pose two sets traffic rules, the static road signs and the dynamic stop signals generated through V2X. After all, we want to make driving easier for humans to avoid potential accidents.

We argue that CAVs can potentially overcome these limitations and thus provide even more efficient and safe driving behaviors in mixed traffic scenarios.
Assisted by vehicle communication, CAVs could learn to show courtesy to improve overall utility~\cite{menendez18iv}.
In this way, not only the intersection management performance can be promoted, but also it might enhance the public acceptance for CAVs.
In this paper we propose a novel method to improve intersection management in mixed traffic,
i.e., the scheduling of vehicles driving through unsignalized intersections,
which represent the majority of all intersections~\cite{10.1145/1860058.1860077}.
We make the following contributions:
\begin{itemize}
    \item We present a centralized intersection management method based on deep reinforcement learning that improves traffic performance at unsignalized intersections through learning cooperation between CAVs and human drivers.
    %that are not directly controlled.
    \item We utilize \textit{return scaling} for training in environments with large imbalance of cumulative rewards at different states.
    \item We present a comprehensive performance comparison for various traffic densities and changing rates of CAVs to demonstrate the potential of our approach.
    %\item Learn cooperation between the IM and human drivers that are not directly controlled by the IM.
    %\item a new formulation of adapting throughput-based reward of each vehicle with equity factor~\cite{9340784}
\end{itemize}
We conduct experimental studies in the traffic simulation environment SUMO~\cite{SUMO2018} and show that our method outperforms two existing intersection management methods on a wide range of traffic densities with varying traffic distributions on the incoming lanes.

\section{Related Work}
\label{sec:related}
Among the first ones to propose an intelligent intersection management system were Dresner and Stone whose reservation-based approach~\cite{dresner2004multiagent, dresner2005multiagent} divides the junction with intersecting trajectories into a grid of tiles. Their autonomous intersection management approach, realized as a centralized controller, applies a first-come-first-served (FCFS) strategy to deal with the requests by CAVs for time slots of the tiles along their trajectories. To accommodate HVs they employ traffic lights and the so-called FCFS-light policy~\cite{IJCAI07-kurt, JAIR08-dresner}. Unlike the obvious benefit of autonomous intersection management, which is designed for pure CAVs, FCFS-light has been shown to provide little or no improvement over today's intersection management methods when less than $90\%$ of the vehicles are automated.
Later, this framework was extended to allow for the centralized intersection management to set the speed profiles of vehicles with cruise control~\cite{au2015autonomous}. %In their setting the driver needs to steer according to the speed regulated by cruise control unit of the vehicle, presenting an additional burden on human drivers and introducing a potential hazard source.
To improve the performance of FCFS-light, Sharon and Stone introduced hybrid autonomous intersection management~\cite{sharon2017protocol}. With this extension, requests of CAVs can be approved regardless of the traffic lights if there are no HVs in the intersecting routes.
%We argue that this behavior is not ideal and might lead to accidents as, on the one hand, it can not account for unexpected behavior of HV and, on the other hand, it will also irritate human drivers due to the differing rule sets for CAVs and HVs.

In general, the methods based on Autonomous Intersection Management provide a relative advantage to CAVs over HVs, which, in our opinion, should be avoided as it might cause the public to repel autonomous vehicles. Furthermore, human drivers will be more sensitive to stopping and waiting than the passengers in CAVs. We therefore suggest that the benefit brought by intersection management and CAVs in general should be evenly shared with human drivers.

Lin \emph{et al.} developed a method similar to the FCFS-light policy~\cite{lin2017autonomous}. It reserves conflicting sections among different routes instead of the grid of tiles. However, this method performs worse than a fixed-time traffic signal controller in traffic with an HV rate above 21\%.
Another first-come-first-served reservation based method has been proposed by Bento \emph{et al.}~\cite{bento2013intelligent}. They suggest to control both CAVs and HVs via speed profiles sent by the intersection management unit. This again places an undesirable burden on human drivers to follow a given speed profile and additionally even requires all HVs to be connected. 

Most of the described approaches make the vehicles roughly follow first-come-first-served order to traverse intersections and correspondingly base the generated speed profiles on this fixed order.
However, as shown by Meng \emph{et al.}~\cite{meng2017analysis}, the performance of an intersection management strategy mainly depends on the passing order of vehicles that it finds. At the same time, the differences caused by trajectory planning algorithms are negligible. Moreover, motion planning approaches like reservation based control often assume that vehicles carry out the planned trajectories precisely. 
In addition, the computation cost grows exponentially with the number of considered vehicles~\cite{meng2017analysis}, which typically leads to simplifying assumptions including linear constraints, no overtaking, no lane changing, constant speed and constant traffic. As shown in our previous work~\cite{9340784}, deep reinforcement learning algorithm with a microscopic simulator can alleviate these problems.
Further, the coordination of the passing order  can mitigate control uncertainties, which makes it more suitable for mixed traffic. Based on this idea, our work is aimed at finding better passing orders, while having vehicles drive based on their own trajectory planning model.

Qian \emph{et al.}~\cite{qian2014priority} assign priorities representing the passing order  to vehicles. While CAVs receive the priority from a central control unit and plan trajectories accordingly, the passing order of HVs is regulated by traffic lights. With high rates of HVs, this potentially results in an inefficient, mostly first-come-first-served control.
Fayazi \emph{et al.}~\cite{fayazi2018mixed} propose to formulate the intersection management problem as a mixed-integer linear program. The intersection management controller assigns times of arrivals to a virtual access area around the junction to CAVs, while HVs are regulated by traffic lights. Since CAVs also need to respect the traffic lights, the performance improvement in mixed traffic is rather limited compared with a fixed-time traffic signal controller.

Most related work on the field so far mainly use fixed-time traffic signal controllers as a baseline for evaluation, which, as shown in~\cite{9340784}, performs sub-optimal compared with a learned adaptive traffic signal controller. Furthermore, up to now most related work considers fixed and relatively low traffic input and requires a high CAV penetration rate to achieve an improved performance. In contrast, we evaluate our proposed method against the state-of-the-art adaptive traffic signal controller in a wide range of dynamic traffic demands and show that the performance gain is available even with a small portion of CAVs in the traffic system.

%DRL implement in traffic control~\cite{kreidieh2018dissipating}

\section{Methods}
\label{sec:methods}
Deep reinforcement learning has shown great potential for solving complex decision making and controlling problems~\cite{mnih2015human,schulman2017proximal}. We model the intersection management task at unsignalized intersections as a Markov Decision Process, where the agent follows a policy $\pi(a \mid s)$ in a specific environment. Based on its state $s_t$ the agent selects an action $a_t\in\mathcal{A}$ according to the policy, transits to a successor state $s_{t+1}$ and receives a reward $r_{t+1}\in\mathbb{R}$. The agent is aimed at maximizing the expectation of the return (discounted cumulative rewards)
\begin{align}
    G(s_t) = \sum_{i > t}\gamma^{i-t-1} r_i,
\end{align}
where $\gamma\in [0,1]$ is the discount factor. We use proximal policy optimization~\cite{schulman2017proximal} to learn the policy $\pi_\theta$ together with the value function $V_{\phi}$.

\begin{figure}[t]
    \centering
    \begin{subfigure}[b]{5.4cm}
    % \begin{subfigure}[b]{0.6\linewidth}
        \centering
        \includegraphics[width=4.9cm]{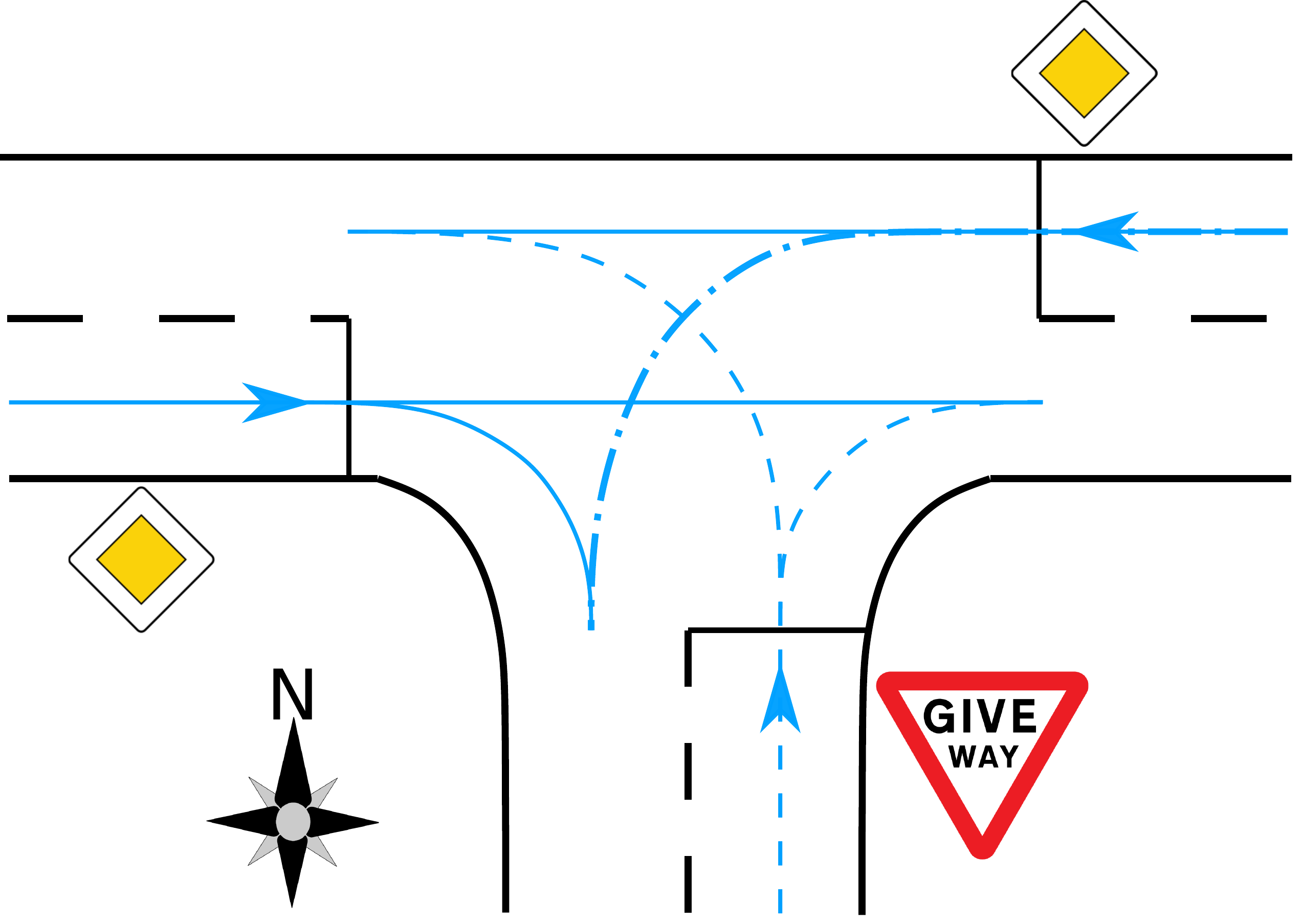}
        \caption{Three-way intersection with six routes.}
        \label{fig:intersection}
    \end{subfigure}\hfill
    \begin{subfigure}[b]{2.8cm}
    % \begin{subfigure}[b]{0.3\linewidth}
        \centering
        \includegraphics[width=2.5cm]{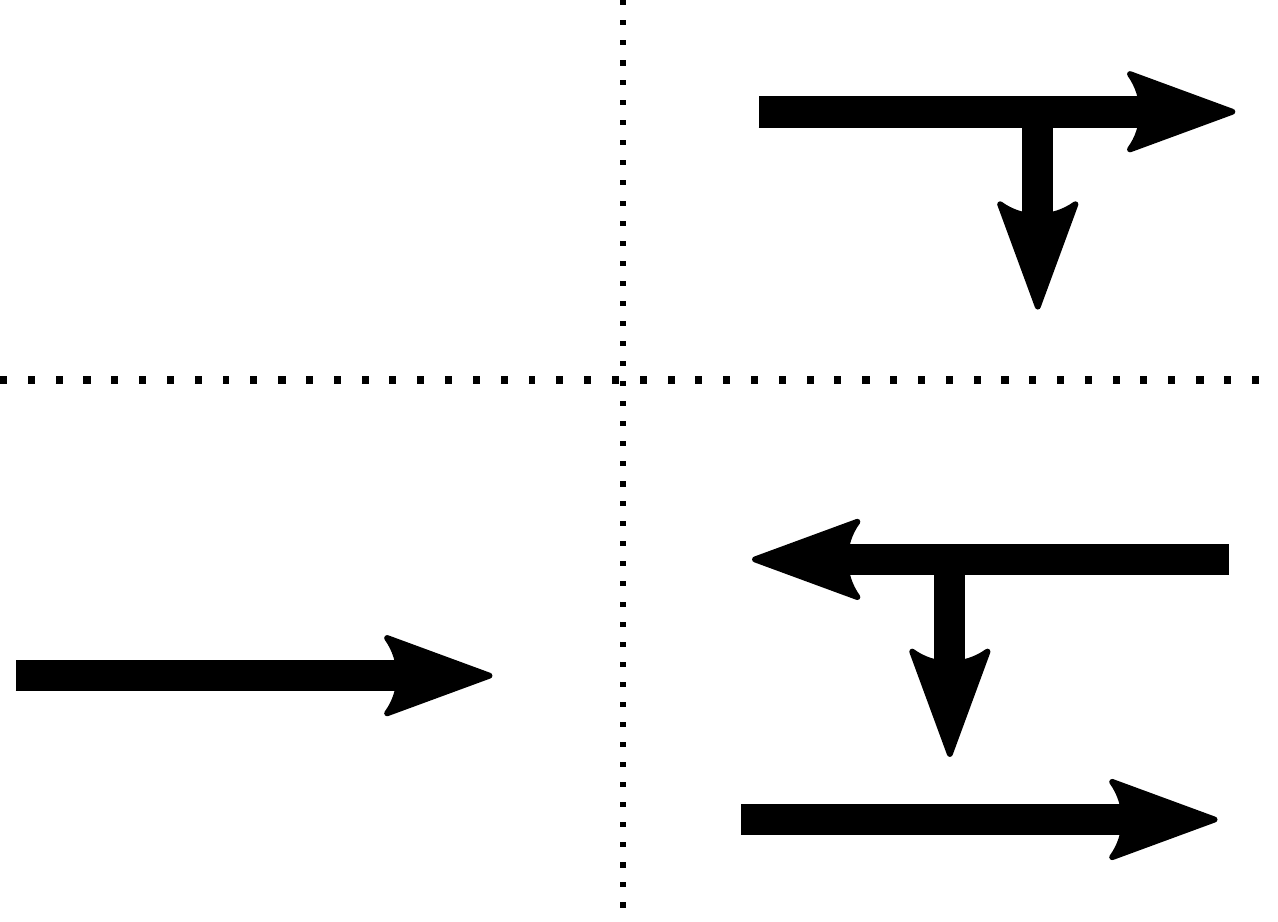}
        \caption{Four actions.}
        \label{fig:actions}
    \end{subfigure}
    \caption{Common regulation of a right-hand traffic three-way intersection (a). The high-priority-routes are W-E, W-S and E-W. The low-priority-routes are S-W and S-E. Route E-S has intersecting routes with higher and lower priority. The proposed set of actions (b) stops CAVs on routes along the indicated directions.}
    \label{fig:intersection_action}
\end{figure}

The method is aimed at training a centralized agent for an intersection that timely stops the CAVs on the routes with higher priority to let the vehicles on conflicting routes with lower priority pass, so that the performance of the whole system is optimized.
Since this is similar to red traffic lights for CAVs on the routes with higher priority, we denote our method as \textit{Courteous Virtual Traffic Signal Control} (CVTSC).
As in our previous work~\cite{9340784}, we evaluate the performance of our method based on both efficiency and equity. In this work, we analyze our proposed approach on the most common type of three-way intersections as illustrated in Fig.~\ref{fig:intersection_action}. By adjusting the state and action representations, our approach could in principle easily be generalized to other intersection layouts, as we show for the real-world intersection in Sec.~\ref{sec:real}. In the following, we introduce the Markov Decision Process formulation in mixed traffic settings.

\subsection{Background}
As we focus on an isolated intersection, we assume that the vehicles can drive freely after they pass the junction and entered the outgoing lanes. Thus the vehicles on the outgoing lanes do not influence the intersection management. However, unlike in our previous work~\cite{9340784}, in which we only considered vehicles in front of the stop lines, we here also take the vehicles into account, which already passed the stop line but not yet entered the outgoing lanes. This is necessary as at unsignalized intersections vehicles very often choose to wait after stop lines and coordination may happen there inside the junction.

In the following we give some definitions of quantities relevant to our approach: 
\begin{itemize}
    %\item Total number of vehicles in the intersection ($N$):
    %    At $t$,
    %    the number of vehicles in the intersection $N_t$ is the total number of vehicles that are within $150\si{\meter}$ to the intersection center (e.g. $150\si{\meter}$) but have not yet entered the outgoing lanes.
    \item Throughput ($N^\text{TP}$):
        The number of vehicles that enter outgoing lanes during step $t$ is denoted $N^{\text{TP}}_t$.
    \item Travel time ($T_\mathit{travel}$):
        For each vehicle passing a junction,
        its travel time is measured as the time period starting from its scheduled spawning time in the simulator (accounting for potential delays caused by traffic jams at the intersection) and ending when it enters an outgoing lane.
        For vehicles not released at the end of an episode, the travel time is counted until the episode ends.
    \item Traffic flow rate ($F$) and Saturation flow rate ($\mathsf{F}_\text{s}$): $F$ represents the number of vehicles (in vehicles per hour $\nicefrac{\text{v}}{\si{\hour}}$) that pass through a point, e.g., an intersection or one lane, in unit time. The term $\mathsf{F}_\text{s}$ is a constant representing the theoretical upper limit for the traffic flow rate.
\end{itemize}

\subsection{Action Space}
\label{sec:action_space}
For the intersection in Fig.~\ref{fig:intersection} we assume that vehicles drive according to the priorities predefined by the road signs, where the diamond indicates priority roads and the triangle indicates yield. Vehicles on the routes with lower priority have to wait until there is enough gap on the conflicting routes with higher priority before passing the junction. Vehicles on the routes with the highest priority, however, can drive freely. Note that in Fig.~\ref{fig:intersection} the route E-S has intersecting routes with higher and lower priority.

To obtain courteous behavior for CAVs on routes with higher priority, without loss of generality, we define a discrete set of four actions \{(), (W-E), (W-E, W-S), (W-E, E-W, E-S)\} as the action space $\mathcal{A}$ in relation to Fig.~\ref{fig:actions}.
The indicated directions show the corresponding routes on which the intersection management unit commands CAVs to halt before the respective stop lines to give priority to vehicles waiting on intersecting routes with lower priority. The action restricting no routes uses the default priorities to manage the intersection. 
We set the duration of each action to 1 second.
%to give the agent more flexibility. 
When a new action $a_t$ is chosen, CAVs on the routes indicated in $a_t$ will receive stopping commands, while the instruction for the routes restricted by $a_{t-1}$ is canceled, if they are not regulated by $a_t$.
If a CAV receives a stopping command while being too close to the stop line, it will continue through the intersection thus ignoring the received command.
Acceleration, collision avoidance and safe distance are managed by the low-level controllers of the individual vehicles (both CAVs and HVs).

Without further adaption our approach can also deal with semi-AVs by assigning them to either the group of CAVs or HVs depending on their level of autonomy. Following our previous argument of acceptability and safety, HVs are not required to change any hardware or driving habits.
%Yet they can still choose to install communication devices to assist at perceiving the courteous intention of the CAVs.

\subsection{State Space}
Due to the restriction of sensors and wireless communication, we assume that the intersection management unit can collect information of vehicles that are within a distance of $150\si{\meter}$ along the road measured from the center of the intersection.
%With the development of perception and C2X communication technologies, a centralized intersection management unit can collect the information of all vehicles within a certain range, i.e. $150\si{\meter}$ in this work. 
We assume that every vehicle's state (position along the road, velocity, time since entering intersection, CAV/HV and its route) is available to the control unit.
Similar to our previous work~\cite{9340784}, the current state $s_t$ of the intersection at time $t$ is given by a vector that contains the structured instantaneous information of vehicles in it. 
%However, the state of traffic lights are removed as they do not exist in this problem. 
The intersection is divided into several lane segments. The capacity of each segment is the maximum amount of vehicles in it during a traffic jam. The states of all vehicles in one segment ordered by their distances to the stop line constitute a part of $s_t$ with a fixed length. Default values are given when fewer vehicles are present than the capacity. The states of all lane segments are concatenated into $s_t$ in a fixed order.

As described in Sec.~\ref{sec:action_space}, only CAVs are controlled by the agent. Every $1$ second a new action should be chosen according to the new state. However, at certain points in time there are no CAVs in the intersection and including these states in training regardless hinders the learning process. We therefore remove states without CAVs from the training data.
%By eliminating such states, the actual state space shrinks, which can accelerate the training process. 
As a result, the influence of actions is not limited to a fixed interval and the duration of one step in the learning process can be any positive integer in seconds. To deal with this variable step length, we employ the method of \textit{adaptive discounting} as proposed by Yan \emph{et al.}~\cite{9340784}.

\subsection{Reward Function}
The common objective of intersection management methods is to improve the efficiency while keeping a certain level of fairness for all vehicles. In this work, we extend the idea of a reward function with equity factor~\cite{9340784}. Instead of using ${T_\mathit{travel}}^\eta$, we propose to use $\eta_\text{a}\cdot{T_\mathit{travel}} + \eta_\text{b}$ as the reward for each released vehicle, where $\eta$, $\eta_\text{a}$ and $\eta_\text{b}$ are equity factors. Due to the flexible step lengths discussed above, the reward of each step $r_t$ is calculated by accumulating discounted rewards generated during step $t$ which might contain up to $k$ environment steps (each one second). I.e., we accumulate the contribution of $N^{\text{TP}}_t$ released vehicles by
\begin{align}
    r_t = \sum_{i=0}^{k-1} \gamma^i \sum_{j=1}^{N^{\text{TP}}_{t\_i}}(\eta_\text{a}\cdot{\tau_j} + \eta_\text{b}),
\end{align}
where $N^{\text{TP}}_{t\_i}$ is the throughput of the $i$th second in step $t$ and the $\tau_j$ are the travel times of the released vehicles.

The values of $\eta_\text{a}$ and $\eta_\text{b}$ are selected as by Yan \emph{et al.}~\cite{9340784} based on two heuristics. First, we favor releasing each vehicle as soon as possible for the purpose of efficiency. The second heuristic aims at equity by considering a traffic situation, where one vehicle waits for saturated traffic flow on an intersecting route with higher priority. Since efficient traffic flow on the high priority route should not be achieved on the expense of accumulating too large waiting time on the single vehicle, we increase the reward contributed by each released vehicle according to its travel time. This linear relation between reward and travel time is more intuitive than the previous exponential formulation. 
Moreover, the additional free variable in this formulation can be used to scale the rewards of single released vehicles to keep them around unity, which is beneficial for hyper-parameter tuning in common deep reinforcement learning setups.

\subsection{Return Scaling}
According to the reward definition, the return $G(s_t)$ is mainly influenced by the throughput and the travel time of released vehicles. Since both of them increase with the traffic input, the scale of $G(s_t)$ could vary from less than $5$ to over $100$ if the state of the intersection changes from nearly empty $s_\text{low}$ to saturated $s_\text{high}$. Consequently, $s_\text{high}$ would have a much larger impact on $\pi_\theta$ and $V_{\phi}$ during the update phase, making the learning process of a policy for light traffic very unstable.

We propose to use \textit{return scaling} to resolve the issues caused by imbalanced return of states, which has shown to be critical for convergence with low traffic volumes in our experiments. In order to reduce the difference between $G(s_\text{low})$ and $G(s_\text{high})$, we scale the cumulative rewards before the update phase with
\begin{align}
    G(s_t) = \rho(s_t)\cdot\sum_{i > t}(\gamma^{\sum_{j=t+2}^{i} k_j}) r_i,
\end{align}
where $k$ is the number of environment steps (each one second) in one step of learning process. The scaling factor $\rho$ is defined as
\begin{align}
    \rho(s_t) = (\nicefrac{N^\text{V}_\text{c}}{n^\text{V}})^{0.2},
\end{align}
where $n^\text{V}$ and $N^\text{V}_\text{c}$ are the current number of vehicles in the intersection and its capacity and $0.2$ is empirically selected.

\section{Experiments}
\label{sec:exp}
\begin{figure*}[t]
    \centering
    \begin{subfigure}[b]{13cm}
    %\begin{subfigure}[b]{0.675\textwidth}
        \centering
        \includegraphics[width=4.2cm]{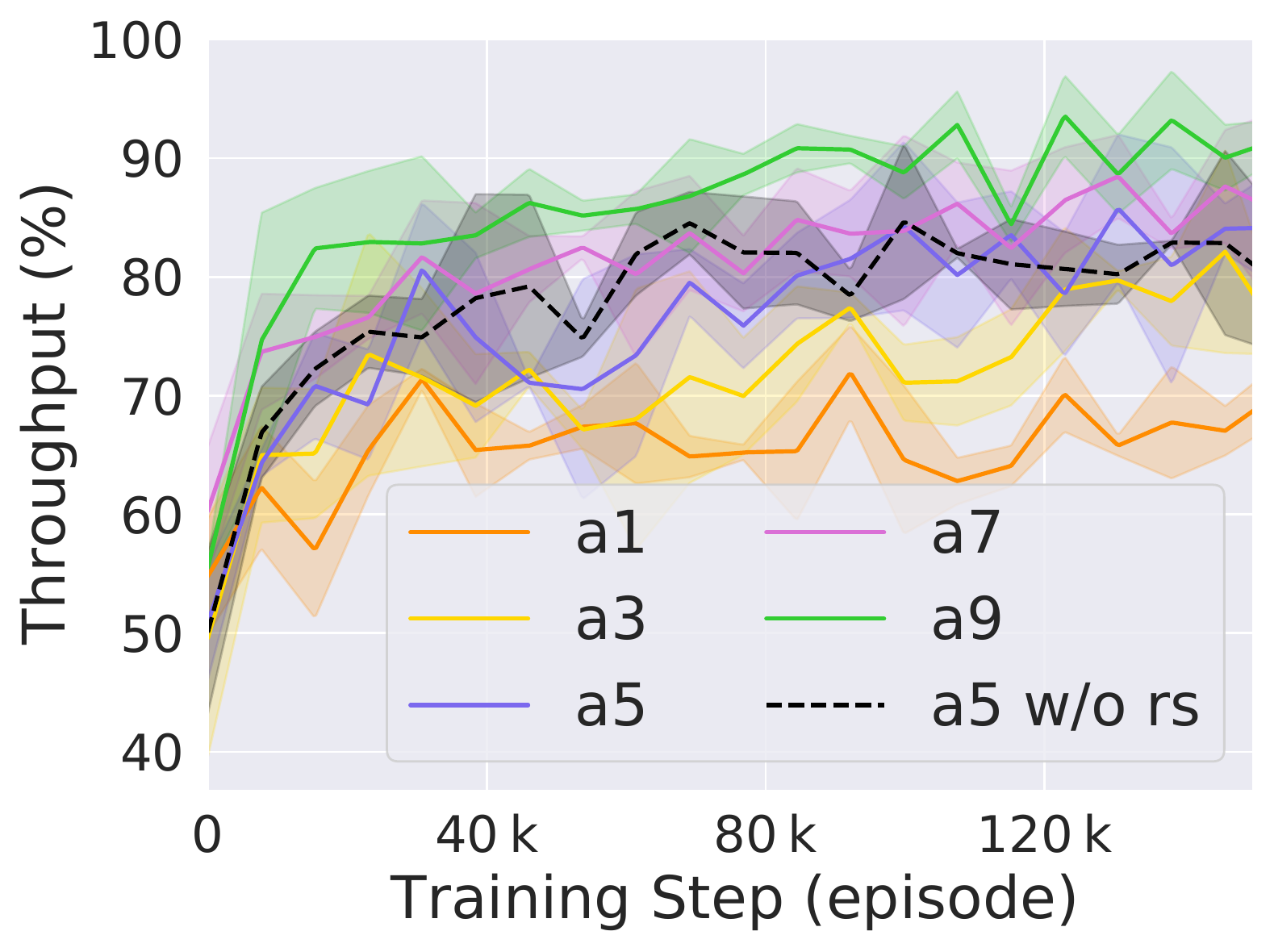}
        \includegraphics[width=4.2cm]{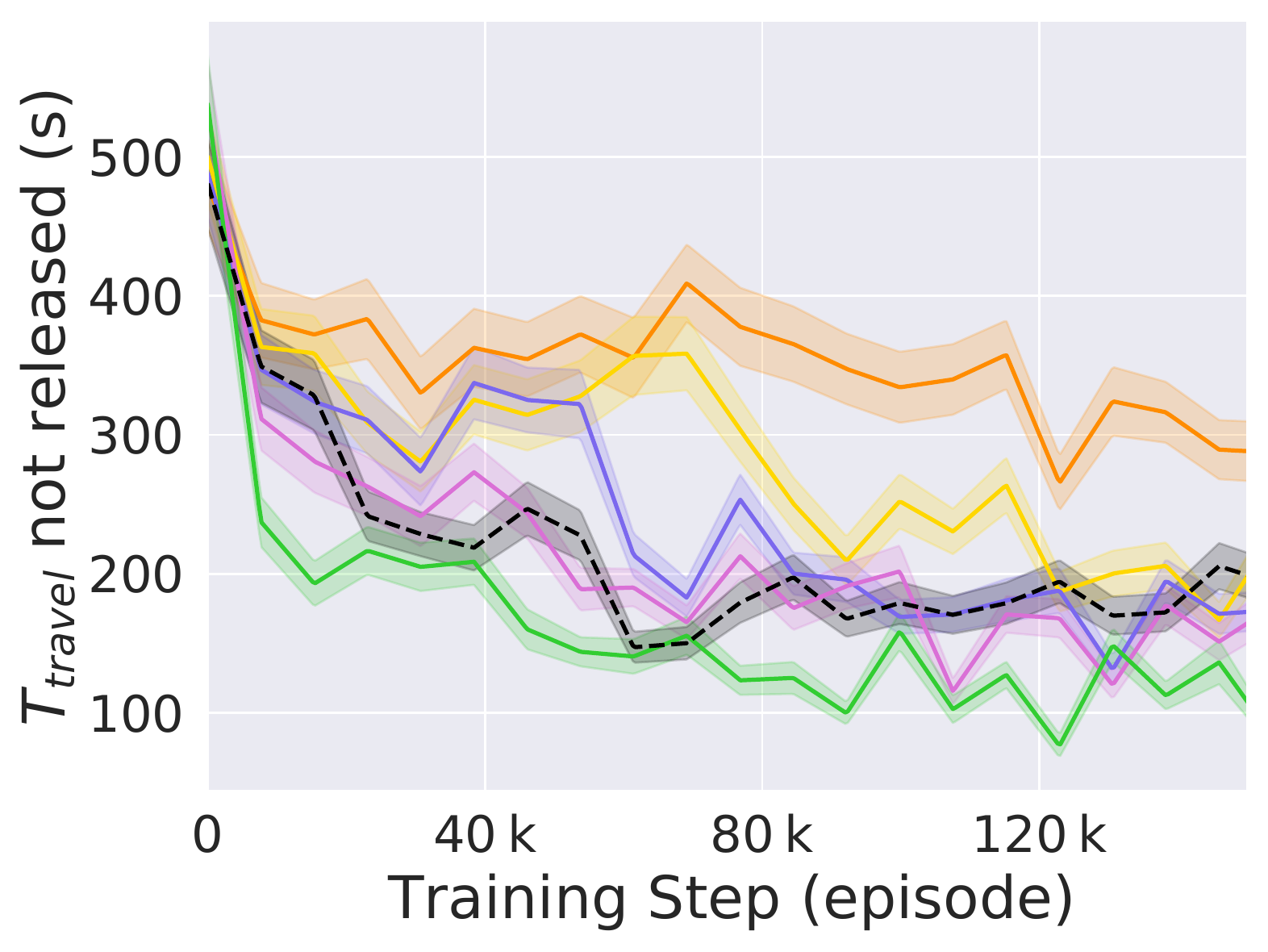}
        \includegraphics[width=4.2cm]{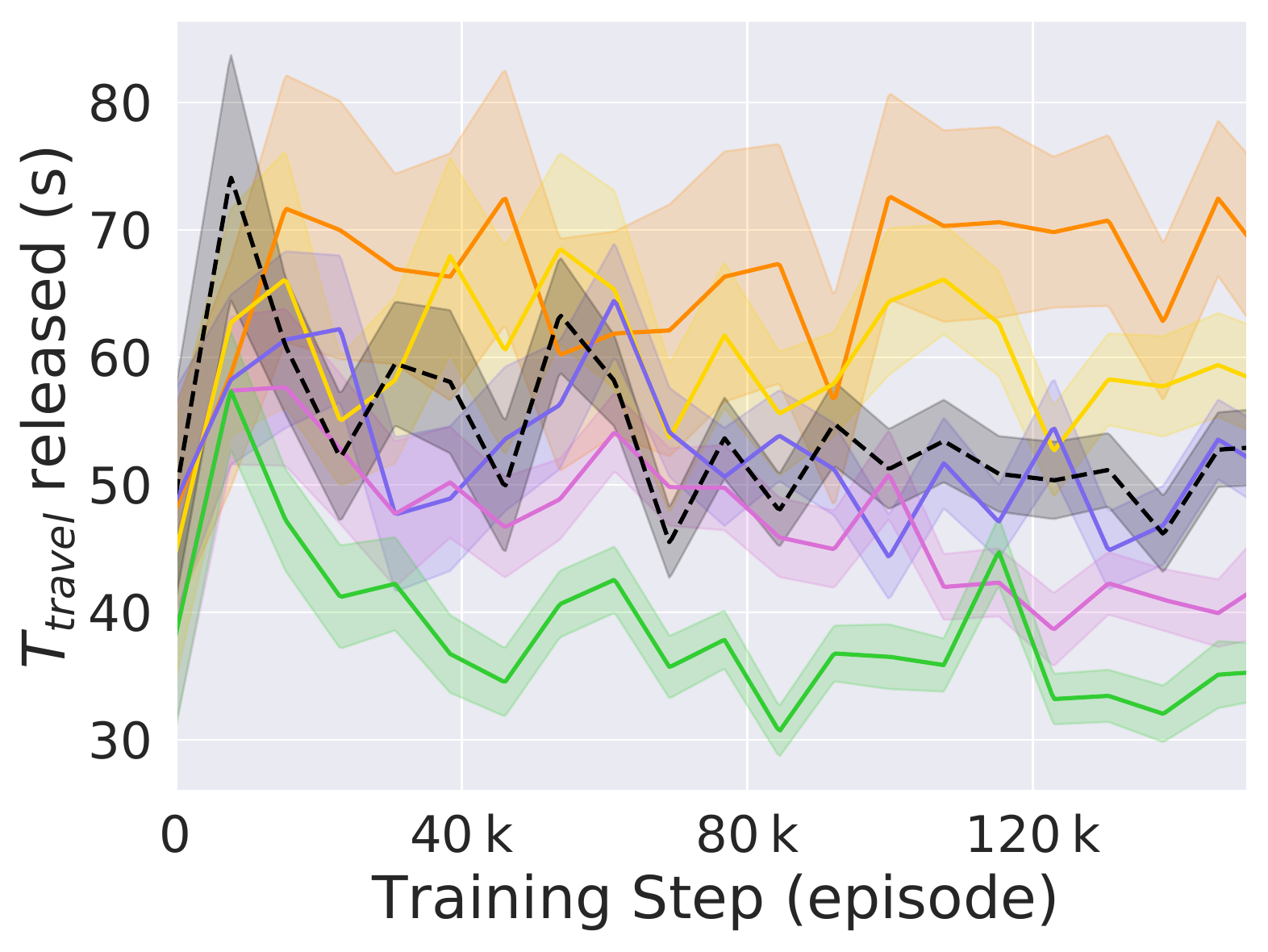}
        \caption{$2\,000\sim3\,000$ $\nicefrac{\text{v}}{\si{\hour}}$}
        \label{fig:curve_high}
    \end{subfigure}\hfill
    \begin{subfigure}[b]{4.5cm}
    %\begin{subfigure}[b]{0.225\textwidth}
        \centering
        \includegraphics[width=4.2cm]{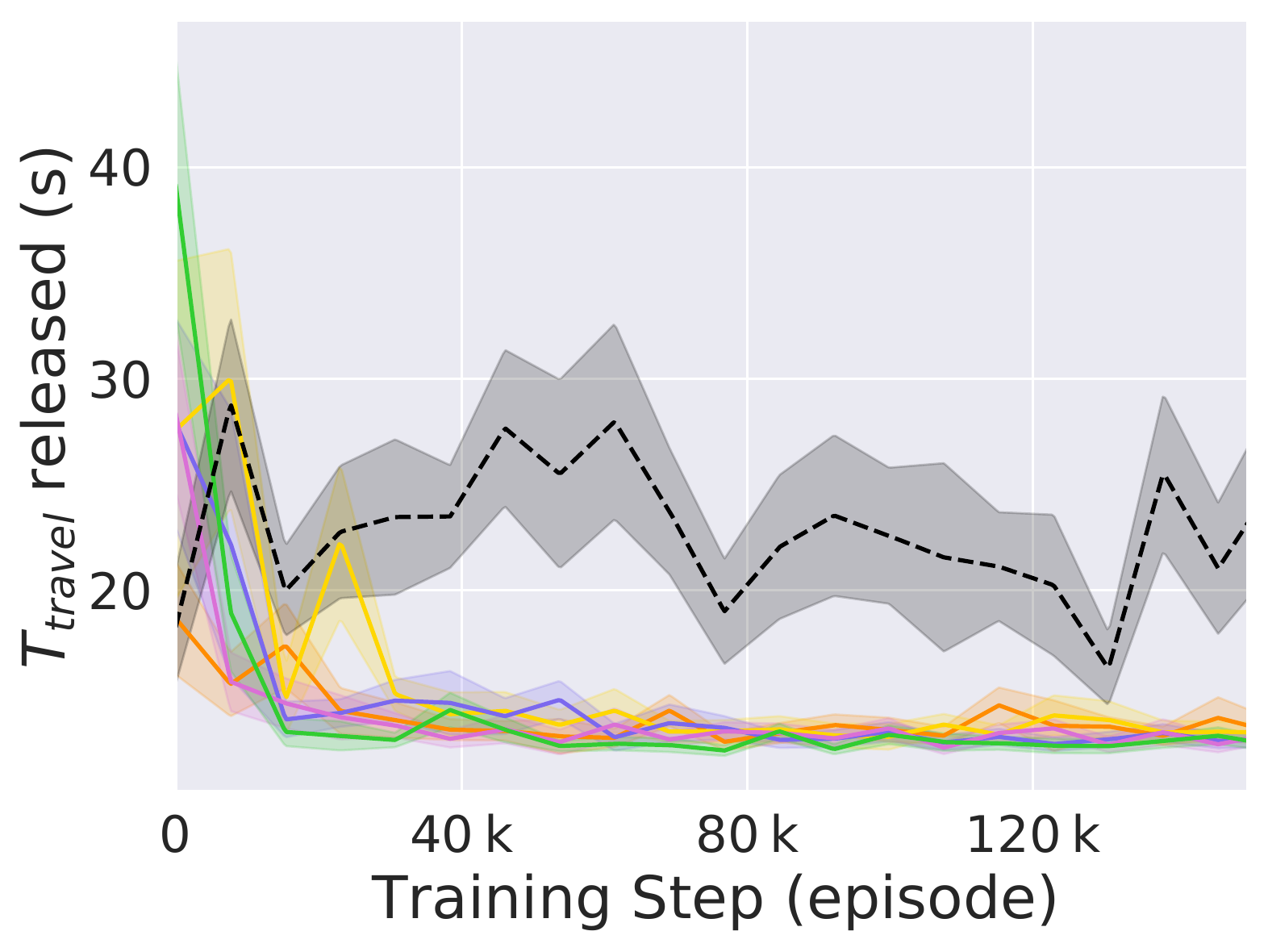}
        \caption{$0\sim1\,000$ $\nicefrac{\text{v}}{\si{\hour}}$}
        \label{fig:curve_low}
    \end{subfigure}
    \caption{Results obtained in evaluation during training for all agents with varying CAV rates in traffic (solid lines) and an ablation study for the usage of the return scaling (dashed black line). 
    The plots show the mean with standard deviation, where the latter is scaled by $\pm\nicefrac{1}{10}$ for the travel times (for clearer visualization), over three non-tuned random seeds. 
    By the end of each episode there are still some vehicles, which have not passed the junction. The travel time for such a vehicle is calculated with $T_\text{episode}-T_\text{spawn}$, where $T_\text{episode}$ is the episode duration and $T_\text{spawn}$ is its scheduled spawning time in the simulator.}
    \label{fig:curves}
\end{figure*}

\subsection{Experimental Setup}
\label{sec:exp-setup}
\begin{figure}[t]
    \centering
    \includegraphics[width=0.25\textwidth]{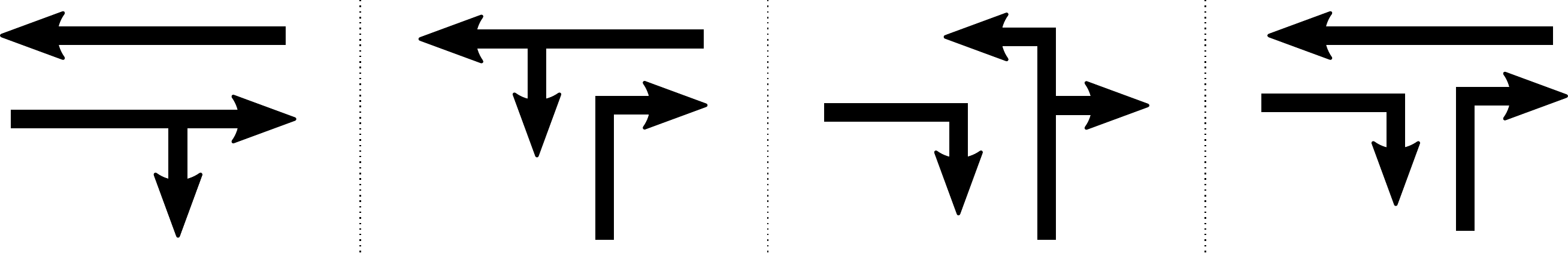}
    \caption{
    Traffic light green phases for the intersection in Fig.~\ref{fig:intersection}.
    }
    \label{fig:actions_tsc}
\end{figure}
We use the open-source traffic simulator SUMO~\cite{SUMO2018} to train and evaluate various intersection management agents.
Besides the simulated traffic episodes we also evaluate our approach on real-world rush hour traffic demand. For all roads we set a speed limit of $50\,\nicefrac{\si{\kilo}\si{\meter}}{\si{\hour}}$.
We compare our approach CVTSC to baselines managing the intersection with \textit{road signs (RS)} defining static priorities for routes and with \textit{traffic lights (TL)} controlled by a deep reinforcement learning agent according to our previous work~\cite{9340784}. Note that this learned controller yields state-of-the-art performance among all traffic signal controllers. Traffic lights will overwrite the priorities defined by road signs. A possible set of green phases for the three-way intersection is shown in Fig.~\ref{fig:actions_tsc}. 

Two fully connected networks $\theta$ and $\phi$ are used as the policy and value function estimators. They both have an input layer of size $343$ and two hidden layers of size 2,048 (ReLU) and 1,024 (ReLU). The output layer is of size $4$ for $\theta$ and $1$ for $\phi$. A grid search was used to select the hyperparameters.
We use $5\mathrm{e}{-6}$ as the learning rate for the Adam optimizer and
$1\mathrm{e}{-3}$ as the coefficient for weight decay.
For proximal policy optimization algorithm,
we use $32$ actors,
the clipping threshold of $\epsilon = 0.001$ and the discount factor of $\gamma = 0.98$.
In each learning step mini-batches of size $100$ are used to update the agents in $8$ epochs. The number of mini-batches in each learning step is, however, variable due to the varying step lengths.
The equity factors $\eta_\text{a}$ and $\eta_\text{b}$ for reward calculation are set to $0.0027$ and $0.946$.

\subsection{Training Setup}
Most current related work has been developed and tested with simplified traffic demand, such as constant traffic input to the intersection. We challenge our approach to train with more dynamic traffic input ranges to cover as many real traffic scenarios as possible. 
For the three-way junction in Fig.~\ref{fig:intersection} the saturation flow rate $\mathsf{F}_s$ of each incoming lane is $1\,670\,\nicefrac{\text{v}}{\si{\hour}}$ and as it is very rare that two non-conflicting routes are simultaneously saturated, we set the traffic demand range to $\left[F_{\min}, F_{\max}\right]=\left[0, 3\,000\right]\,\nicefrac{\text{v}}{\si{\hour}}$.

%traffic flow upper bound of  can be calculated as $2\cdot\mathsf{F}_s$, where the . As 

We train our agents online on simulated traffic episodes with a duration of $1\,200\si{\second}$. First, the total traffic input $F_\text{begin}$ is randomly sampled in $\left[F_{\min}, F_{\max}\right]$. Then $F_\text{end}$ is sampled uniformly within 
$[\max(F_{\min}, F_{\text{begin}}-1\,500), 
  \min(F_{\max}, F_{\text{begin}}+1\,500)]$. 
After that the beginning and ending traffic flow for each route is randomly sampled from an uniform distribution, such that they sum up to $F_\text{begin}$ and $F_\text{end}$, respectively. Finally, the traffic flow during the episode is generated by linear interpolation between these two values for each route.

%As the change from $0$ to $100\%$ of CAV penetration rate in the traffic system will presumably take decades, we believe it is practical to train different agents for increasing rate of CAVs and employ the specific agent according to the current CAV rate in the real traffic. 
%To update the intersection management unit for a higher CAV penetration rate, people only need to change the neural network. 
We train five agents (\textit{a1, a3, a5, a7, a9}), each corresponding to a fixed CAV rate of $\left[10, 30, 50, 70, 90\right]\%$, corresponding to the expected increasing CAV rates in the future traffic. In the following sections, we first show how these agents can optimize the intersection management performance for their respective CAV rate. Then we cross-evaluate them on settings corresponding to different CAV penetration rates.
% and show that they also perform very well in the traffic with different CAV rates than their native during training.

\subsection{Evaluation during Training}
To monitor the learning process the performance is evaluated for traffic input of different ranges: $\left[0, 1\,000\right]$, $\left[500, 1\,500\right]$, $\left[1\,000, 2\,000\right]$, $\left[1\,500, 2\,500\right]$, $\left[2\,000, 3\,000\right]$. The generation of traffic demand is analogous to that of training episodes except that the total traffic inputs at the beginning $F_{\text{begin}}$ and end $F_{\text{end}}$ are sampled independently in the five given ranges.

The plots in Fig.~\ref{fig:curves} show the performance of agents trained with
different CAV rates and present an ablation study for the usage of the \textit{return scaling}.
%are the evaluation results of the ablation study for analysing the contribution of  and return scaling in our algorithm. 
The agent \textit{a5~w/o~rs} is trained with a CAV rate of $50\%$ without using return scaling. We analyze the throughput in percentage of released vehicles among all spawned vehicles, the travel time of released and not released vehicles at the highest traffic density level and the travel time of released vehicles at the lowest level. The calculated travel time is the mean among all released or not released vehicles during three evaluation episodes.
We analyze the throughput and travel times instead of the accumulated reward as they give us a better estimate of the overall performance. The variance of the travel times is of particular interest as it is a good indicator for the equity. Large variances correspond to some vehicles with long waiting times at the intersection.
%Since the traffic episodes are randomly sampled for each evaluation phase, 
%The overall traffic input and its split of vehicles over the different routes are sampled randomly. 
%Consequently, the travel time variance of the released vehicles could be very big especially for the bad policies.

As illustrated in Fig.~\ref{fig:curve_high}, CVTSC with higher CAV rate leads to more throughput, more efficient clearance (lower average $T_{\mathit{travel}}$) of the intersection and more fairness (shown by lower standard deviation of $T_{\mathit{travel}}$) to all the vehicles. As expected, from Fig.~\ref{fig:curve_low} and the travel time plots of Fig.~\ref{fig:curve_high}, we observe that the agent without return scaling fails to learn an efficient policy for light traffic, although its performance is similar to that of \textit{a5} in heavy traffic.
We plan to conduct further investigation on return scaling, in particular whether it is applicable to a broader class of problems or can be replaced with other methods like $\gamma$-tuning.

\subsection{Evaluation on Simulated Traffic Demand}
We first test our agents with simulated traffic episodes, each with a duration of one hour. 
For each of the five traffic demand levels described above, we first create $50$ traffic episodes with spawning time of each vehicle following the procedure to that for evaluation during training. Then we generate five sets of mixed traffic episodes with different CAV rates by randomly setting each vehicle as CAV or HV according to the penetration rate.
Note that the baseline methods \textit{road sign (RS)} and \textit{traffic light (TL)} do not distinguish between CAV and HV.
Following this setup, we test both baselines and our trained agents with identical number of vehicles and same spawning times.
In the following, the five agents are first tested with their corresponding CAV rates $\left[10, 30, 50, 70, 90\right]\%$ to evaluate their performance against the baseline methods. Then we analyze the performance of each agent on all the five traffic settings.

\subsubsection{Performance of Intersection}
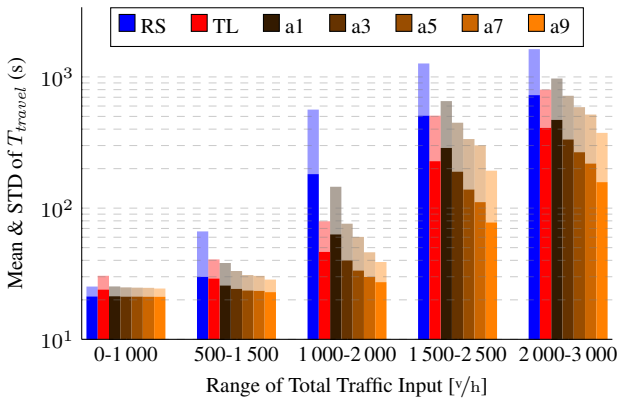
\begin{figure}[t]
	\centering
	\setlength{\fboxrule}{0pt}
	\fbox{
		\scalefont{0.8}
		\begin{tikzpicture}[
		%\centering[
		every axis/.style={
		    ymode=log, log basis y=10, ybar stacked,
			axis on top,
			height=6cm, width=\columnwidth,
			bar width=0.15cm,
			ymajorgrids, tick align=inside,
			yminorgrids, tick align=inside,
			grid style={dashed,gray,opacity=0.4},
			%minor grid style={dotted,gray!50!white}
			enlarge y limits={value=.1,upper},
			ymin=10, ymax=2000,
			ytick={10,100,1000},
			minor ytick={20,30,40,50,60,70,80,90,200,300,400,500,600,700,800,900},
			axis x line*=bottom,
			axis y line*=left,
			%y axis line style={opacity=0},
			tickwidth=0pt,
			subtickwidth=0pt,
			enlarge x limits=true,
			legend style={
				at={(0.5,1.)},
				anchor=north,
				legend columns=-1,
				/tikz/every even column/.append style={column sep=0.5cm}
			},
			ylabel={Mean \& STD of $T_\mathit{travel}$ (s)},
			xlabel={Range of Total Traffic Input [$\nicefrac{\text{v}}{\si{\hour}}$]},
			symbolic x coords={
				0-1\,000,500-1\,500,1\,000-2\,000,1\,500-2\,500,
				2\,000-3\,000},
			xtick=data,
		}
		]
		
		% TS
		\begin{axis}[bar shift=-0.45cm, hide axis]
		\addplot [draw=none, fill=blue] coordinates
		{(0-1\,000, 21.24) (500-1\,500, 29.96) (1\,000-2\,000, 181.72) (1\,500-2\,500, 507.09) (2\,000-3\,000, 726.52)};
		\addplot [draw=none, fill=blue, fill opacity=0.4] coordinates
		{(0-1\,000, 4.02) (500-1\,500, 36.52) (1\,000-2\,000, 381.99) (1\,500-2\,500, 759.44) (2\,000-3\,000, 901.04)};
		\end{axis}
	    % a1
		\begin{axis}[bar shift=-0.15cm, hide axis]
		\addplot [draw=none, fill=black!80!orange] coordinates
		{(0-1\,000, 21.32) (500-1\,500, 25.76) (1\,000-2\,000, 63.16) (1\,500-2\,500, 287.76) (2\,000-3\,000, 471.01)};
		\addplot [draw=none, fill=black!80!orange, fill opacity=0.4] coordinates
		{(0-1\,000, 3.99) (500-1\,500, 12.43) (1\,000-2\,000, 82.38) (1\,500-2\,500, 366.58) (2\,000-3\,000, 501.44)};
		\end{axis}
		% a3
		\begin{axis}[bar shift=0cm, hide axis]
		\addplot [draw=none, fill=black!60!orange] coordinates
		{(0-1\,000, 21.19) (500-1\,500, 24.26) (1\,000-2\,000, 40.01) (1\,500-2\,500, 189.96) (2\,000-3\,000, 334.18)};
		\addplot [draw=none, fill=black!60!orange, fill opacity=0.4] coordinates
		{(0-1\,000, 3.70) (500-1\,500, 8.92) (1\,000-2\,000, 36.1) (1\,500-2\,500, 257.84) (2\,000-3\,000, 384.46)};
		\end{axis}
		% a5
		\begin{axis}[bar shift=0.15cm, hide axis]
		\addplot [draw=none, fill=black!40!orange] coordinates
		{(0-1\,000, 21.20) (500-1\,500, 23.60) (1\,000-2\,000, 33.50) (1\,500-2\,500, 138.83) (2\,000-3\,000, 266.96)};
		\addplot [draw=none, fill=black!40!orange, fill opacity=0.4] coordinates
		{(0-1\,000, 3.58) (500-1\,500, 7.29) (1\,000-2\,000, 26.88) (1\,500-2\,500, 197.46) (2\,000-3\,000, 321.35)};
		\end{axis}
		% a7
		\begin{axis}[bar shift=0.3cm, hide axis]
		\addplot [draw=none, fill=black!20!orange] coordinates
		{(0-1\,000, 21.15) (500-1\,500, 23.44) (1\,000-2\,000, 29.90) (1\,500-2\,500, 111.15) (2\,000-3\,000, 218.99)};
		\addplot [draw=none, fill=black!20!orange, fill opacity=0.4] coordinates
		{(0-1\,000, 3.55) (500-1\,500, 7.03) (1\,000-2\,000, 16.19) (1\,500-2\,500, 189.97) (2\,000-3\,000, 297.81)};
		\end{axis}
		% a9
		\begin{axis}[bar shift=0.45cm, hide axis]
		\addplot [draw=none, fill=orange] coordinates
		{(0-1\,000, 21.13) (500-1\,500, 22.90) (1\,000-2\,000, 27.29) (1\,500-2\,500, 77.88) (2\,000-3\,000, 157.84)};
		\addplot [draw=none, fill=orange, fill opacity=0.4] coordinates
		{(0-1\,000, 3.27) (500-1\,500, 5.65) (1\,000-2\,000, 11.57) (1\,500-2\,500, 115.18) (2\,000-3\,000, 216.89)};
		\end{axis}
		% TL
		\begin{axis}[bar shift=-0.3cm, legend style={/tikz/every even column/.append style={column sep=0.25cm}}]%, legend style={draw=none}]
		\addplot [draw=none, fill=blue] coordinates
		{(0-1\,000, 1e-6) (500-1\,500, 1e-6) (1\,000-2\,000, 1e-6) (1\,500-2\,500, 1e-6) (2\,000-3\,000, 1e-6)};
		\addplot [draw=none, fill=red] coordinates
		{(0-1\,000, 23.97) (500-1\,500, 29.10) (1\,000-2\,000, 46.42) (1\,500-2\,500, 228.02) (2\,000-3\,000, 409.32)};
		\addplot [draw=none, fill=black!80!orange] coordinates
		{(0-1\,000, 1e-6) (500-1\,500, 1e-6) (1\,000-2\,000, 1e-6) (1\,500-2\,500, 1e-6) (2\,000-3\,000, 1e-6)};
		\addplot [draw=none, fill=black!60!orange] coordinates
		{(0-1\,000, 1e-6) (500-1\,500, 1e-6) (1\,000-2\,000, 1e-6) (1\,500-2\,500, 1e-6) (2\,000-3\,000, 1e-6)};
		\addplot [draw=none, fill=black!40!orange] coordinates
		{(0-1\,000, 1e-6) (500-1\,500, 1e-6) (1\,000-2\,000, 1e-6) (1\,500-2\,500, 1e-6) (2\,000-3\,000, 1e-6)};
		\addplot [draw=none, fill=black!20!orange] coordinates
		{(0-1\,000, 1e-6) (500-1\,500, 1e-6) (1\,000-2\,000, 1e-6) (1\,500-2\,500, 1e-6) (2\,000-3\,000, 1e-6)};
		\addplot [draw=none, fill=orange] coordinates
		{(0-1\,000, 1e-6) (500-1\,500, 1e-6) (1\,000-2\,000, 1e-6) (1\,500-2\,500, 1e-6) (2\,000-3\,000, 1e-6)};
		\addplot [draw=none, fill=red, fill opacity=0.4] coordinates
		{(0-1\,000, 6.54) (500-1\,500, 11.57) (1\,000-2\,000, 32.99) (1\,500-2\,500, 278.49) (2\,000-3\,000, 390.97)};

		\legend{RS, TL, a1, a3, a5, a7, a9}
		\end{axis}
		\end{tikzpicture}
	}
	\caption{%\protect\raggedright
	Performance comparison of our CVTSC with baselines RS and TL in traffics with different CAV rates.
	%on $250$ one-hour simulated demand episodes ($50$ from each of the $5$ traffic densities). 
	For each controller with each traffic density, the mean (opaque bars) and positive standard deviation (translucent bars) of $T_\mathit{travel}$ are calculated over all vehicles (including released and not released) of $50$ simulated traffic episodes.
    Each CVTSC agent is trained and evaluated in traffics with its corresponding CAV rate.
	}

    \label{fig:tt_intersection}
\end{figure}

\begin{table}[t]
    \scalefont{0.9}
    \begin{tabular}{rccccccc}
		\toprule
        	\multirow{2}{*}{\shortstack{Traffic Input\\($\nicefrac{\text{v}}{\si{\hour}}$)}} & \multicolumn{7}{c}{Throughput $(\%)$} \\
		\cmidrule{2-8}
			& RS & a1 & a3 & a5 & a7 & a9 & TL\\
		\midrule
	 	    $\hspace{.06in}0\sim1\,000$ & $99.4$ & $99.4$ & $99.4$ & $99.4$ & $99.4$ & $99.4$ & $99.4$ \\
	        $500\sim1\,500$ & $99.2$ & $99.3$ & $99.3$ & $\mathbf{99.4}$ & $99.3$ & $\mathbf{99.4}$ & $99.2$ \\
	        $1\,000\sim2\,000$ & $91.1$ & $97.7$ & $98.6$ & $99.0$ & $99.1$ & $\mathbf{99.2}$ & $98.5$ \\
	        $1\,500\sim2\,500$ & $72.2$ & $85.3$ & $90.6$ & $93.5$ & $94.7$ & $\mathbf{96.8}$ & $88.5$ \\
	        $2\,000\sim3\,000$ & $59.8$ & $74.6$ & $82.1$ & $85.8$ & $88.5$ & $\mathbf{91.9}$ & $77.9$ \\
		\bottomrule
    \end{tabular}
    \caption{
        Throughput ($\%$) of considered methods in Fig.~\ref{fig:tt_intersection}.
    }
	\label{tab:tp_intersection}
\end{table}

The performance is shown in Fig.~\ref{fig:tt_intersection} and Table~\ref{tab:tp_intersection}. For all the tested traffic density levels, our CVTSC agents can improve the performance of the unsignalized intersection. Not only more vehicles are released during the same period, but also the mean and standard deviation of their travel times are reduced. The higher the CAV rate is, the better our approach performs. The performance gain of CVTSC on the lowest traffic density is not obvious, because nearly no vehicles have to stop at the junction. When there is little traffic, employing \textit{TL} can cause unnecessary stopping due to the transition phase (amber or red lights). In heavier traffic over $1\,500\nicefrac{\text{v}}{\si{\hour}}$ \textit{TL} outperforms \textit{a1} by a little margin. However, it is outperformed by CVTSC when $30\%$ or more vehicles are CAVs.

\subsubsection{Performance of Vehicle Groups}
\pgfplotstableread{
x       y       y-min   y-max
RS      21.61   1.88    5.63
a1      23.9    3.66    20.74
a3      22.81   2.75    12.16
a5      22.12   2.24    6.92
a7      21.63   1.89    4.17
a9      21.58   1.83    3.87
}{\hvmaintt}
\pgfplotstableread{
x       y       y-min   y-max
RS      21.61   1.88    5.63
a1      34.66   12.01   26.65
a3      24.56   3.81    14.91
a5      23.27   2.82    8.34
a7      22.32   2.2     4.97
a9      22.26   2.13    4.54
}{\avmaintt}
\pgfplotstableread{
x       y       y-min   y-max
RS      162.34  120.4   392.75
a1      63.39   31.53   62.73
a3      39.28   13.32   21.68
a5      33.73   9.4     15.73
a7      33.18   8.69    12.68  
a9      27.14   4.7     8.27
}{\vsidett}

\pgfplotstableread{
x       y
RS      98.87
a1    98.65
a3    99.15
a5    99.24
a7    99.35
a9    99.13
}{\hvmaintp}
\pgfplotstableread{
x       y
RS      98.87
a1    98.27
a3    99.01
a5    99.18
a7    99.23
a9    99.25
}{\avmaintp}
\pgfplotstableread{
x       y
RS      74.27
a1    95.61
a3    97.64
a5    98.34
a7    98.66
a9    99.08
}{\vsidetp}
% only for the line style at ylabels
\pgfplotstableread{
x       y
RS      2000
}{\plotlabel}

\begin{figure}[t]
	\centering
	\scalefont{0.8}
	\begin{tikzpicture}
	\begin{axis} [axis on top,
			height=6cm,
			ymode=log,
			width=0.9*\columnwidth,
			bar width=0.15cm,
			%ymajorgrids, tick align=inside,
			%yminorgrids, tick align=inside,
			%grid style={dashed,gray,opacity=0.4},
			enlarge y limits={value=.1,upper},
			ymin=1e1, ymax=1e3,
			%max space between ticks=20,
			%ytick={10, 100, 1000},
			%minor ytick={20,30,40,50,60,70,80,90,75,125},
			axis x line*=bottom,
			axis y line*=left,
			%tickwidth=0pt,
			xtick style={draw=none},
			%subtickwidth=0pt,
			enlarge x limits=true,
			ylabel={\ref{plt:ttplot} $T_\mathit{travel}$ released [$s$]},
			symbolic x coords={RS, a1, a3, a5, a7, a9},xtick=data,legend style={only marks, at={(0.95,0.8)},anchor=north east}, legend cell align={left},legend entries={Main road HV, Main road CAV, Side road all}]
	\addplot+[ mark=triangle,mark options={scale=1}, xshift=0.15cm,legend image post style={xshift=-0.15cm}, color=red, error bars/.cd, y dir=both, y explicit]
	table[x=x,y=y,y error plus expr=\thisrow{y-max},y error minus expr=\thisrow{y-min}] {\hvmaintt};
	\addplot+[ mark=star,mark options={scale=1}, xshift=0cm, color=blue, error bars/.cd, y dir=both, y explicit]
	table[x=x,y=y,y error plus expr=\thisrow{y-max},y error minus expr=\thisrow{y-min}] {\avmaintt};
	\addplot+[ mark=diamond,mark options={scale=1}, xshift=-0.15cm,legend image post style={xshift=0.15cm}, color=orange, error bars/.cd, y dir=both, y explicit]
	table[x=x,y=y,y error plus expr=\thisrow{y-max},y error minus expr=\thisrow{y-min}] {\vsidett};
	\addplot+[ mark=none, color=black]
	table[x=x,y=y] {\plotlabel};\label{plt:ttplot}
	\end{axis} 
	
	\begin{axis}[axis on top,
			height=6cm, width=0.9*\columnwidth,
			bar width=0.15cm,
			enlarge y limits={value=.1,upper},
			ymin=70, ymax=100,
			axis x line*=bottom,
			axis y line*=right,
			xtick style={draw=none},
			subtickwidth=0pt,
			enlarge x limits=true,
			ylabel={\ref{plt:tpplot} Throughput [\%]},
			symbolic x coords={RS, a1, a3, a5, a7, a9},
			xtick=\empty,
			x axis line style={draw=none}]
	\addplot+[ mark=triangle,mark options={scale=1,solid}, xshift=0.15cm, color=red, dashed]
	table[x=x,y=y] {\hvmaintp};
	\addplot+[ mark=star,mark options={scale=1,solid}, color=blue, dashed]
	table[x=x,y=y] {\avmaintp};
	\addplot+[ mark=diamond,mark options={scale=1,solid}, xshift=-0.15cm, color=orange, dashed]
	table[x=x,y=y] {\vsidetp};
	\addplot+[ mark=none, color=black, dashed]
	table[x=x,y=y] {\plotlabel};\label{plt:tpplot}
	\end{axis}
	
    \end{tikzpicture}
    \caption{Performance comparison of different vehicle groups at traffic demand $1\,000\sim2\,000$ $\nicefrac{\text{v}}{\si{\hour}}$. The plotted travel times show the median, lower quartile and higher quartile over all released vehicles among all evaluated episodes.
    The plotted throughput is the percentage of released vehicles among all spawned vehicles throughout all episodes.  }
    \label{fig:performance_groups}
\end{figure}
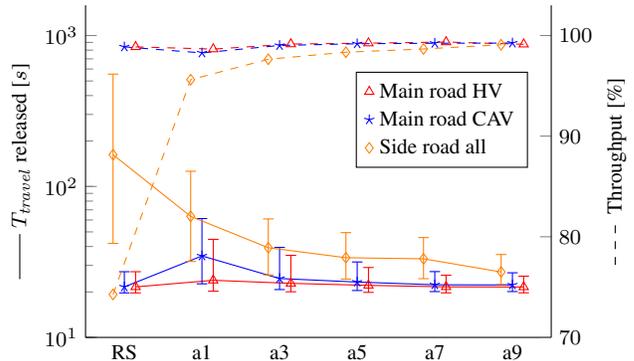
\begin{table*}[t]
    \centering
    \scalefont{0.9}
    \begin{tabular}{crrrrrrccccc}
		\toprule
        	%\multirow{2}{*}{\shortstack{CAV Percentage\\(\%)}} & %\multirow{2}{*}{\shortstack{Traffic Input\\(v/h)}}
        	\multicolumn{2}{c}{Traffic Input} & \multicolumn{5}{c}{Average $T_\mathit{travel}$ [$\si{\second}$]} & \multicolumn{5}{c}{Throughput $[\%]$} \\
		\cmidrule(lr){1-2} \cmidrule(lr){3-7} \cmidrule(lr){8-12}
			CAV Rate & Input Flow [$\nicefrac{\text{v}}{\si{\hour}}$] & a1 & a3 & a5 & a7 & a9 & a1 & a3 & a5 & a7 & a9 \\
		\midrule
		    \multirow{4}{*}{\shortstack{$10\%$}}
		    & $500\sim1\,500$ & $25.8$ & $\mathbf{25.7}$ & $25.8$ & $26.3$ & $27.5$ & $99.3$ & $99.3$ & $99.3$ & $99.3$ & $99.3$\\
		    & $1\,000\sim2\,000$ & $\mathbf{63.2}$ & $69.9$ & $80.8$ & $102.1$ & $131.4$ & $\mathbf{97.7}$ & $97.4$ & $96.9$ & $95.8$ & $94.2$\\
		    & $1\,500\sim2\,500$ & $\mathbf{287.8}$ & $299.3$ & $339.0$ & $364.2$ & $432.3$ & $\mathbf{85.3}$ & $84.7$ & $82.1$ & $80.6$ & $77.1$\\
		    & $2\,000\sim3\,000$ & $\mathbf{471.0}$ & $482.5$ & $517.8$ & $554.4$ & $610.1$ & $\mathbf{74.6}$ & $73.9$ & $72.0$ & $69.3$ & $65.9$\\
		\cmidrule(lr){3-7} \cmidrule(lr){8-12}
		    \multirow{4}{*}{\shortstack{$30\%$}}
		    & $500\sim1\,500$ & $24.7$ & $\mathbf{24.3}$ & $24.4$ & $24.9$ & $24.8$ & $99.3$ & $99.3$ & $99.3$ & $99.3$ & $99.3$\\
		    & $1\,000\sim2\,000$ & $42.1$ & $\mathbf{40.0}$ & $43.4$ & $49.4$ & $58.7$ & $98.5$ & $\mathbf{98.6}$ & $\mathbf{98.6}$ & $98.2$ & $98.0$\\
		    & $1\,500\sim2\,500$ & $213.4$ & $\mathbf{190.0}$ & $204.2$ & $237.9$ & $274.1$ & $89.5$ & $\mathbf{90.6}$ & $90.0$ & $88.1$ & $85.9$\\
		    & $2\,000\sim3\,000$ & $367.3$ & $\mathbf{334.2}$ & $347.0$ & $411.7$ & $430.6$ & $80.3$ & $\mathbf{82.1}$ & $81.4$ & $77.5$ & $76.6$\\
		\cmidrule(lr){3-7} \cmidrule(lr){8-12}
		    \multirow{4}{*}{\shortstack{$50\%$}}
		    & $500\sim1\,500$ & $24.1$ & $\mathbf{23.6}$ & $\mathbf{23.6}$ & $23.9$ & $23.9$ & $99.3$ & $99.3$ & $\mathbf{99.4}$ & $99.3$ & $99.3$\\
		    & $1\,000\sim2\,000$ & $36.0$ & $33.7$ & $\mathbf{33.5}$ & $35.6$ & $38.9$ & $98.9$ & $\mathbf{99.0}$ & $\mathbf{99.0}$ & $98.9$ & $98.7$\\
		    & $1\,500\sim2\,500$ & $191.0$ & $145.5$ & $\mathbf{138.8}$ & $159.7$ & $174.7$ & $90.6$ & $93.2$ & $\mathbf{93.5}$ & $92.3$ & $91.8$\\
		    & $2\,000\sim3\,000$ & $346.9$ & $269.0$ & $\mathbf{267.0}$ & $308.6$ & $313.4$ & $81.4$ & $\mathbf{85.9}$ & $85.8$ & $83.5$ & $83.3$\\
		\cmidrule(lr){3-7} \cmidrule(lr){8-12}
		    \multirow{4}{*}{\shortstack{$70\%$}}
		    & $500\sim1\,500$ & $23.6$ & $23.2$ & $\mathbf{23.1}$ & $23.4$ & $23.3$ & $\mathbf{99.4}$ & $\mathbf{99.4}$ & $\mathbf{99.4}$ & $99.3$ & $99.3$\\
		    & $1\,000\sim2\,000$ & $32.6$ & $29.8$ & $\mathbf{28.6}$ & $29.9$ & $30.0$ & $99.0$ & $\mathbf{99.1}$ & $\mathbf{99.1}$ & $\mathbf{99.1}$ & $\mathbf{99.1}$\\
		    & $1\,500\sim2\,500$ & $176.3$ & $120.7$ & $\mathbf{101.1}$ & $111.2$ & $112.1$ & $91.2$ & $94.4$ & $\mathbf{95.6}$ & $94.7$ & $95.0$\\
		    & $2\,000\sim3\,000$ & $323.2$ & $234.2$ & $\mathbf{203.5}$ & $219.0$ & $217.0$ & $82.6$ & $87.3$ & $\mathbf{89.3}$ & $88.5$ & $88.5$\\
		\cmidrule(lr){3-7} \cmidrule(lr){8-12}
		    \multirow{4}{*}{\shortstack{$90\%$}}
		    & $500\sim1\,500$ & $23.1$ & $22.8$ & $\mathbf{22.6}$ & $23.1$ & $22.9$ & $\mathbf{99.4}$ & $\mathbf{99.4}$ & $\mathbf{99.4}$ & $99.3$ & $\mathbf{99.4}$\\
		    & $1\,000\sim2\,000$ & $30.3$ & $27.9$ & $\mathbf{26.7}$ & $27.4$ & $27.3$ & $99.0$ & $\mathbf{99.2}$ & $\mathbf{99.2}$ & $\mathbf{99.2}$ & $\mathbf{99.2}$\\
		    & $1\,500\sim2\,500$ & $164.8$ & $105.5$ & $77.0$ & $\mathbf{76.5}$ & $77.9$ & $91.8$ & $95.2$ & $\mathbf{96.8}$ & $96.7$ & $\mathbf{96.8}$\\
		    & $2\,000\sim3\,000$ & $311.5$ & $192.0$ & $161.6$ & $\mathbf{154.5}$ & $157.8$ & $83.2$ & $90.0$ & $91.9$ & $\mathbf{92.2}$ & $91.9$\\
		\bottomrule
    \end{tabular}
    \caption{
        Performance comparison of different agents with different traffic input settings.
        For each agent with each traffic setting, the average $T_\mathit{travel}$ is calculated over all vehicles (including released and not released) of $50$ simulated traffic episodes.
    }
	\label{tab:agents_performance}
\end{table*}

In contrast to the relative advantage of CAVs over HVs suggested by the methods based on autonomous intersection management, our CVTSC tends to share the performance gain evenly between the two types of vehicles.
Fig.~\ref{fig:performance_groups} shows how CVTSC can increase the intersection management performance while keeping the balance between different vehicle categories. Since the actions are executed only for CAVs on the main road,
we divide vehicles on the main road into \textit{Main~road~CAV} and \textit{Main~road~HV}
and assign all vehicles on the side road to a third group \textit{Side~road~all}. As illustrated, the performance gain against \textit{RS} is mainly caused by the improvement of the traffic on the side road. With only $10\%$ CAVs the throughput of the side road traffic is increased from $74.3\%$ to $95.6\%$ and the median travel time is decreased by $61\%$. As a necessary side effect, the courteous behavior adds about $13\,\si{\second}$ to the average travel time of CAVs on the main road and slows down some HVs following the CAVs consequently. However, the average travel time of Main~road~HV and the throughput of both vehicle groups on the main road are nearly not influenced. With growing rate of CAVs in traffic, the performance of the traffic on the side road continues to be improved while the initial disadvantage for the main road is compensated.
%of the main road becomes equal to or even a bit better than the performance without CVTSC.

\subsubsection{Comparison of Agents}
To cross-evaluate their performance on other traffic settings than their natives, we further test each agent (a1 to a9) on the five different CAV rates on $50$ simulated episodes on each of the five traffic densities. 
Since CVTSC brings nearly no measurable difference for the lowest traffic density, only the results for the other four traffic densities are listed in Table~\ref{tab:agents_performance}.

We observe that all trained CVTSC agents outperform \textit{RS} in any mixed traffic setting.
Furthermore, two significant patterns can be observed in the results.
First, for each CAV rate the agents trained with similar rate values are among the best, as expected.
Second, as the CAV rate increases the performance of all agents is continuously improved.
Interestingly, \textit{a5}, the one trained with CAV rate of $50\%$, outperforms or performs equally well as \textit{a7} and \textit{a9} even in settings where CAVs are the majority. We suppose this is because \textit{a5} during training is exposed to more diverse traffic situations, especially ones with fewer CAVs in the intersection. As shown in Fig.~\ref{fig:tt_intersection} and Fig.~\ref{fig:performance_groups}, the margin of the performance gain decreases with increased CAV rate. Even though \textit{a7} and \textit{a9} can handle highly automated traffic better than \textit{a5}, the performance gain is so small that it can not compensate the performance loss when occasionally more HVs drive in the intersection.

%These results can provide a guidance for the employment of the CVTSC agents in the real traffic. Before the local level of CAV ownership grows up to $50\%$, agents \textit{a1} or \textit{a3} can be employed according to the rate value. With more CAVs in the traffic \textit{a5} should be adopted.

\subsection{Evaluation on Real-world Traffic Demand}
\label{sec:real}
\begin{figure}[t]
    \centering
    \includegraphics[width=0.25\textwidth]{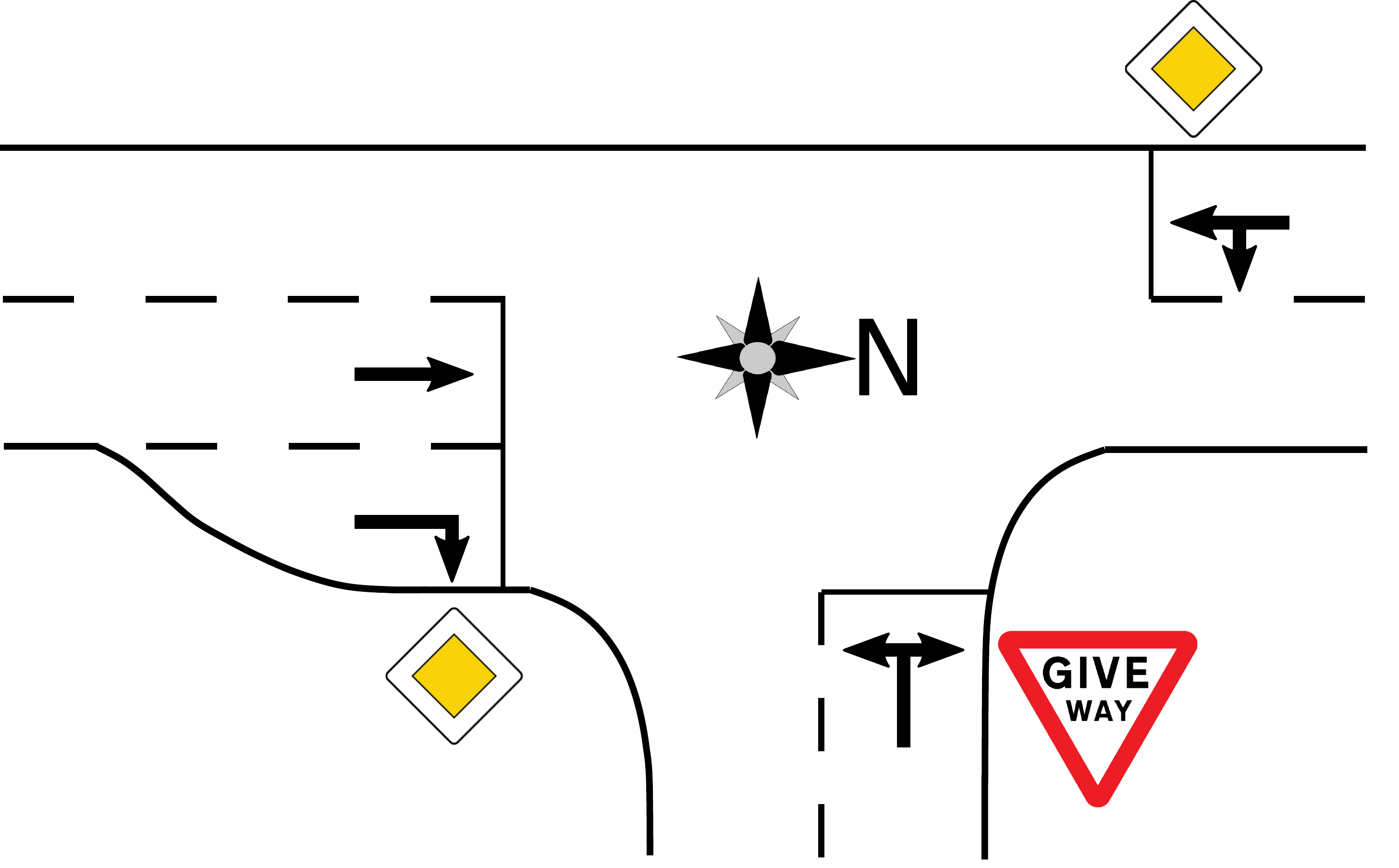}
    \caption{
    Intersection of Tullastrasse and Hans-Bunte-Strasse in Freiburg, Germany.
    }    
    \label{fig:intersection_tulla}
\end{figure}

\begin{figure}[t]
    \centering
    \includegraphics[width=0.48\textwidth]{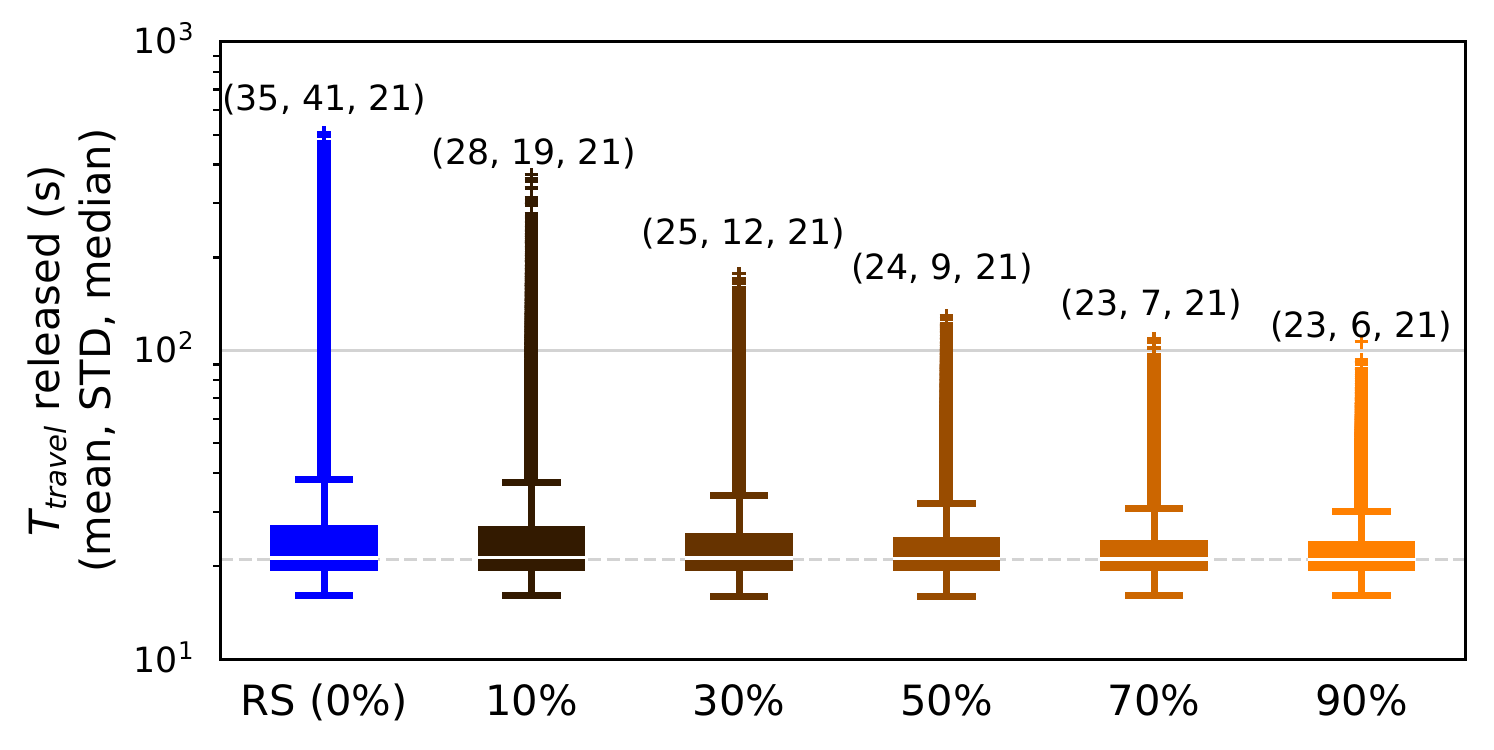}
	\caption{Box plot of travel times with different CAV rates over all released vehicles in the simulation based on the real-world intersection of Fig.~\ref{fig:intersection_tulla}. The whiskers extend $1.5\cdot\text{IQR}$ (interquartile range) from the upper and lower quartiles.
    }    
    \label{fig:real_perform}
\end{figure}

To further evaluate CVTSC in more realistic traffic situations, we conduct additional tests with real-world traffic demand recorded at an intersection in Freiburg, Germany, which is sketched in Fig.~\ref{fig:intersection_tulla}. Unlike the intersection above, one part of the main road (Tullastrasse) forks before the stop line. After adjusting the state representation and the intersection structure in the simulator we trained two new agents \textit{a3} and \textit{a5} and employ them in the test. The traffic demand, listed in Table~\ref{tab:traffic_tulla}, was manually recorded on October 19, 2017 by the traffic department of Freiburg. The total traffic input was about $1\,000\sim1\,500\,\nicefrac{\text{v}}{\si{\hour}}$ with roughly $20\%$ on the side road.

Fig.~\ref{fig:real_perform} shows box plots of the travel times of released vehicles controlled by \textit{RS} and CVTSC agents in traffic scenarios with different CAV rates. The agent \textit{a3} is employed for $10\%$ and $30\%$ automated traffic, while \textit{a5} is employed for the other three. In all scenarios over $99.7\%$ of all vehicles traverse the intersection. Our method continuously improves the traffic flow with increasing rate of CAVs in traffic. We notice that the median of travel times in all scenarios stay similar, which means the performance gain comes mainly from the vehicles with long travel times on the side road. CVTSC agents manage to release them faster without delaying the traffic on the main road.

\begin{table*}[t]
    \centering
    \scalefont{0.9}
    \begin{tabular}{crrrrrrrrrrrrrrrrr}
		\toprule
        	\multirow{2}{*}{\shortstack{Direction}} & 
        	\multicolumn{17}{c}{Traffic Input (Number of Vehicles every $15\,\si{\minute}$)} \\
        	\cmidrule(lr){2-18}
        	& 7:15 & 7:30 & 7:45 & 8:00 & 8:15 & 8:30 & 8:45 & 9:00 &  & 16:15 & 16:30 & 16:45 & 17:00 & 17:15 & 17:30 & 17:45 & 18:00 \\
		\midrule
		N-S & 55 & 63 & 101 & 80 & 98 & 85 & 60 & 111 &     & 102 & 104 & 79 & 97 & 148 & 122 & 104 & 67\\
		N-E & 44 & 29 & 38 & 44 & 32 & 44 & 28 & 31 &    & 32 & 44 & 26 & 28 & 32 & 37 & 38 & 19\\
		S-N & 71 & 76 & 96 & 111 & 78 & 86 & 80 & 65 &     & 105 & 88 & 119 & 116 & 112 & 86 & 100 & 108\\
		S-E & 35 & 41 & 32 & 53 & 68 & 42 & 52 & 43 &     & 29 & 32 & 29 & 36 & 33 & 30 & 29 & 27\\
		E-N & 11 & 26 & 29 & 20 & 40 & 29 & 20 & 22 &     & 58 & 48 & 56 & 35 & 55 & 50 & 47 & 35\\
		E-S & 16 & 25 & 51 & 26 & 31 & 21 & 32 & 22 &    & 53 & 32 & 43 & 23 & 32 & 19 & 31 & 25\\
		\bottomrule
    \end{tabular}
    \caption{
        Traffic in rush hours on the morning and afternoon of October 19, 2017 at the intersection of Fig.~\ref{fig:intersection_tulla}.
    }
	\label{tab:traffic_tulla}
\end{table*}

\section{Conclusion}
\label{sec:conclusion}
In this paper we present a novel approach to managing mixed traffic of autonomous and human-driven vehicles at unsignalized intersections using deep reinforcement learning. Our proposed method CVTSC creates courtesy behavior similar to human drivers for autonomous vehicles in order to optimize the overall traffic flow at intersections. 
Furthermore, we propose to use return scaling to reduce the imbalance of cumulative rewards at different states and to stabilize training.
We validate the effectiveness of CVTSC using simulated and real-world traffic data and  
show that CVTSC improves the performance of unsignalized intersections continuously with increasing percentage of autonomous vehicles. For more than $10\%$ of autonomous vehicles it also outperforms the state-of-the-art adaptive traffic signal controller without the need for traffic lights.
Besides the benefit in intersection performance, our method does not require a change of the current driving habits of humans. Moreover it is fault-tolerant, since the method is an add-on to the existing traffic rules and thus the intersection will still be fully functional even if the intersection management unit fails.
%, which is also practical for maintenance or update. 
Last but not least, our method can be easily adopted to different intersection topologies.
%As active cooperation is necessary for human drivers, we propose to employ CVTSC at small intersections, which are nowadays governed by traffic signs. Within this scope our method can be easily adopted to different intersection topologies.

\addtolength{\textheight}{-12cm}   % This command serves to balance the column lengths

%%%%%%%%%%%%%%%%%%%%%%%%%%%%%%%%%%%%%%%%%%%%%%%%%%%%%%%%%%%%%%%%%%%%%%%%%%%%%%%%

%%%%%%%%%%%%%%%%%%%%%%%%%%%%%%%%%%%%%%%%%%%%%%%%%%%%%%%%%%%%%%%%%%%%%%%%%%%%%%%%

% \section*{ACKNOWLEDGMENT}

%%%%%%%%%%%%%%%%%%%%%%%%%%%%%%%%%%%%%%%%%%%%%%%%%%%%%%%%%%%%%%%%%%%%%%%%%%%%%%%%
\bibliographystyle{IEEEtran}
\bibliography{yan21}

\begin{thebibliography}{10}
\providecommand{\url}[1]{#1}
\csname url@rmstyle\endcsname
\providecommand{\newblock}{\relax}
\providecommand{\bibinfo}[2]{#2}
\providecommand\BIBentrySTDinterwordspacing{\spaceskip=0pt\relax}
\providecommand\BIBentryALTinterwordstretchfactor{4}
\providecommand\BIBentryALTinterwordspacing{\spaceskip=\fontdimen2\font plus
\BIBentryALTinterwordstretchfactor\fontdimen3\font minus
  \fontdimen4\font\relax}
\providecommand\BIBforeignlanguage[2]{{%
\expandafter\ifx\csname l@#1\endcsname\relax
\typeout{** WARNING: IEEEtran.bst: No hyphenation pattern has been}%
\typeout{** loaded for the language `#1'. Using the pattern for}%
\typeout{** the default language instead.}%
\else
\language=\csname l@#1\endcsname
\fi
#2}}

\bibitem{wu2015influence}
L.~Wu, Y.~Ci, J.~Chu, and H.~Zhang, ``The influence of intersections on fuel
  consumption in urban arterial road traffic: A single vehicle test in harbin,
  china,'' \emph{PloS one}, vol.~10, no.~9, p. e0137477, 2015.

\bibitem{MUTCD2009}
U.~F.~H. Administration, \emph{{Manual on Uniform Traffic Control Devices}},
  2009, [Online; accessed 06-Mar-2021].

\bibitem{varaiya2013max}
P.~Varaiya, ``The max-pressure controller for arbitrary networks of signalized
  intersections,'' in \emph{Advances in Dynamic Network Modeling in Complex
  Transportation Systems}.\hskip 1em plus 0.5em minus 0.4em\relax Springer,
  2013, pp. 27--66.

\bibitem{9340784}
\BIBentryALTinterwordspacing
S.~{Yan}, J.~{Zhang}, D.~{Büscher}, and W.~{Burgard}, ``Efficiency and equity
  are both essential: A generalized traffic signal controller with deep
  reinforcement learning,'' in \emph{Proc.~of the IEEE/RSJ International
  Conference on Intelligent Robots and Systems (IROS)}, 2020, pp. 5526--5533.
  [Online]. Available:
  \url{http://ais.informatik.uni-freiburg.de/publications/papers/yan20iros.pdf}
\BIBentrySTDinterwordspacing

\bibitem{litman2020autonomous}
T.~Litman, ``{Autonomous vehicle implementation predictions: Implications for
  transport planning},'' \url{https://www.vtpi.org/avip.pdf}, 2021, [Online;
  accessed 06-Mar-2021].

\bibitem{namazi2019intelligent}
E.~Namazi, J.~Li, and C.~Lu, ``Intelligent intersection management systems
  considering autonomous vehicles: A systematic literature review,'' \emph{IEEE
  Access}, vol.~7, pp. 91\,946--91\,965, 2019.

\bibitem{ulbrich2015structuring}
S.~Ulbrich, S.~Grossjohann, C.~Appelt, K.~Homeier, J.~Rieken, and M.~Maurer,
  ``Structuring cooperative behavior planning implementations for automated
  driving,'' in \emph{2015 IEEE 18th International Conference on Intelligent
  Transportation Systems}.\hskip 1em plus 0.5em minus 0.4em\relax IEEE, 2015,
  pp. 2159--2165.

\bibitem{menendez18iv}
\BIBentryALTinterwordspacing
C.~Men\'{e}ndez-Romero, M.~Sezer, F.~Winkler, C.~Dornhege, and W.~Burgard,
  ``Courtesy behavior for highly automated vehicles on highway interchanges,''
  in \emph{IEEE Intelligent Vehicles Symposium (IV)}, 2018, pp. 943--948.
  [Online]. Available:
  \url{http://ais.informatik.uni-freiburg.de/publications/papers/menendez18iv.pdf}
\BIBentrySTDinterwordspacing

\bibitem{10.1145/1860058.1860077}
M.~Ferreira, R.~Fernandes, H.~Concei\c{c}\~{a}o, W.~Viriyasitavat, and O.~K.
  Tonguz, ``Self-organized traffic control,'' in \emph{Proc.~of the Seventh ACM
  International Workshop on VehiculAr InterNETworking}, ser. VANET '10, New
  York, NY, USA, 2010, p. 85–90.

\bibitem{SUMO2018}
P.~A. {Lopez}, M.~{Behrisch}, L.~{Bieker-Walz}, J.~{Erdmann}, Y.~{Flötteröd},
  R.~{Hilbrich}, L.~{Lücken}, J.~{Rummel}, P.~{Wagner}, and E.~{Wiessner},
  ``Microscopic traffic simulation using sumo,'' in \emph{Proc.~of the IEEE
  International Conference on Intelligent Transportation Systems (ITSC)}, 2018,
  pp. 2575--2582.

\bibitem{dresner2004multiagent}
K.~Dresner and P.~Stone, ``Multiagent traffic management: A reservation-based
  intersection control mechanism,'' in \emph{Autonomous Agents and Multiagent
  Systems, International Joint Conference on}, vol.~3.\hskip 1em plus 0.5em
  minus 0.4em\relax IEEE Computer Society, 2004, pp. 530--537.

\bibitem{dresner2005multiagent}
------, ``Multiagent traffic management: An improved intersection control
  mechanism,'' in \emph{Proceedings of the fourth international joint
  conference on Autonomous agents and multiagent systems}, 2005, pp. 471--477.

\bibitem{IJCAI07-kurt}
------, ``Sharing the road: Autonomous vehicles meet human drivers,'' in
  \emph{The 20th International Joint Conference on Artificial Intelligence},
  January 2007, pp. 1263--68.

\bibitem{JAIR08-dresner}
------, ``A multiagent approach to autonomous intersection management,''
  \emph{Journal of Artificial Intelligence Research}, vol.~31, pp. 591--656,
  March 2008.

\bibitem{au2015autonomous}
T.-C. Au, S.~Zhang, and P.~Stone, ``Autonomous intersection management for
  semi-autonomous vehicles,'' \emph{Handbook of transportation}, pp. 88--104,
  2015.

\bibitem{sharon2017protocol}
G.~Sharon and P.~Stone, ``A protocol for mixed autonomous and human-operated
  vehicles at intersections,'' in \emph{International Conference on Autonomous
  Agents and Multiagent Systems}.\hskip 1em plus 0.5em minus 0.4em\relax
  Springer, 2017, pp. 151--167.

\bibitem{lin2017autonomous}
P.~Lin, J.~Liu, P.~J. Jin, and B.~Ran, ``Autonomous vehicle-intersection
  coordination method in a connected vehicle environment,'' \emph{IEEE
  Intelligent Transportation Systems Magazine}, vol.~9, no.~4, pp. 37--47,
  2017.

\bibitem{bento2013intelligent}
L.~C. Bento, R.~Parafita, S.~Santos, and U.~Nunes, ``Intelligent traffic
  management at intersections: Legacy mode for vehicles not equipped with v2v
  and v2i communications,'' in \emph{Proc.~of the IEEE International Conference
  on Intelligent Transportation Systems (ITSC)}.\hskip 1em plus 0.5em minus
  0.4em\relax IEEE, 2013, pp. 726--731.

\bibitem{meng2017analysis}
Y.~Meng, L.~Li, F.-Y. Wang, K.~Li, and Z.~Li, ``Analysis of cooperative driving
  strategies for nonsignalized intersections,'' \emph{IEEE Transactions on
  Vehicular Technology}, vol.~67, no.~4, pp. 2900--2911, 2017.

\bibitem{qian2014priority}
X.~Qian, J.~Gregoire, F.~Moutarde, and A.~De~La~Fortelle, ``Priority-based
  coordination of autonomous and legacy vehicles at intersection,'' in
  \emph{Proc.~of the IEEE International Conference on Intelligent
  Transportation Systems (ITSC)}.\hskip 1em plus 0.5em minus 0.4em\relax IEEE,
  2014, pp. 1166--1171.

\bibitem{fayazi2018mixed}
S.~A. Fayazi and A.~Vahidi, ``Mixed-integer linear programming for optimal
  scheduling of autonomous vehicle intersection crossing,'' \emph{IEEE
  Transactions on Intelligent Vehicles}, vol.~3, no.~3, pp. 287--299, 2018.

\bibitem{mnih2015human}
V.~Mnih, K.~Kavukcuoglu, D.~Silver, A.~A. Rusu, J.~Veness, M.~G. Bellemare,
  A.~Graves, M.~Riedmiller, A.~K. Fidjeland, G.~Ostrovski, \emph{et~al.},
  ``Human-level control through deep reinforcement learning,'' \emph{Nature},
  vol. 518, no. 7540, pp. 529--533, 2015.

\bibitem{schulman2017proximal}
J.~Schulman, F.~Wolski, P.~Dhariwal, A.~Radford, and O.~Klimov, ``Proximal
  policy optimization algorithms,'' \emph{arXiv preprint arXiv:1707.06347},
  2017.

\end{thebibliography}

\end{document}